\documentclass{article}

\usepackage{amsmath,amsfonts,bm}
\usepackage{xspace}

\newcommand{\DMH}{\texttt{DM-HARD-8}\xspace}
\newcommand{\RND}{\texttt{RND}\xspace}
\newcommand{\RNDonAP}{\texttt{RNDonAP}\xspace}
\newcommand{\NGU}{\texttt{NGU}\xspace}
\newcommand{\MEME}{\texttt{MEME}\xspace}
\newcommand{\BYOLE}{\texttt{BYOL-Explore}\xspace}
\newcommand{\VMPO}{\texttt{VMPO}\xspace}
\newcommand{\EMM}{\texttt{EMM}\xspace}
\newcommand{\DETOCS}{\texttt{RECODE}\xspace}
\newcommand{\CASM}{\texttt{CASM}\xspace}
\newcommand{\CASnotM}{\texttt{CASnotM}\xspace}
\newcommand{\AP}{\texttt{AP}\xspace}
\newcommand{\dm}{d_{\text{ema}}}
\newcommand{\irc}{{n_0}}

\usepackage{mathtools}

\def\eqref#1{equation~\ref{#1}}

\def\1{\bm{1}}

\DeclareMathAlphabet{\mathsfit}{\encodingdefault}{\sfdefault}{m}{sl}
\SetMathAlphabet{\mathsfit}{bold}{\encodingdefault}{\sfdefault}{bx}{n}

\DeclareMathOperator*{\argmin}{arg\,min}

\usepackage{microtype}
\usepackage{graphicx}
\usepackage{booktabs} 

\usepackage{hyperref}

\usepackage[accepted]{icml2023}

\usepackage{amsmath}
\usepackage{amssymb}
\usepackage{mathtools}
\usepackage{amsthm}

\usepackage[capitalize,noabbrev]{cleveref}

\theoremstyle{plain}

\theoremstyle{definition}

\theoremstyle{remark}

\usepackage[textsize=tiny]{todonotes}

\usepackage{url}
\usepackage{setspace}
\usepackage{wrapfig}
\usepackage{dsfont}
\usepackage{subcaption}

\usepackage{xspace}

\icmltitlerunning{Unlocking the Power of Representations in Long-term Novelty-based Exploration}

\begin{document}

\twocolumn[
\icmltitle{Unlocking the Power of Representations\\* in Long-term Novelty-based Exploration}

\icmlsetsymbol{equal}{*}

\begin{icmlauthorlist}
\icmlauthor{Alaa Saade}{equal,yyy}
\icmlauthor{Steven Kapturowski}{equal,yyy}
\icmlauthor{Daniele Calandriello}{equal,yyy}

\icmlauthor{Charles Blundell}{yyy}
\icmlauthor{Pablo Sprechmann}{yyy}
\icmlauthor{Leopoldo Sarra}{zzz}
\icmlauthor{Oliver Groth}{yyy}
\icmlauthor{Michal Valko}{yyy}
\icmlauthor{Bilal Piot}{yyy}
\end{icmlauthorlist}

\icmlaffiliation{yyy}{DeepMind}
\icmlaffiliation{zzz}{Department of Physics, Friedrich-Alexander Universität Erlangen-Nürnberg, work done while interning at DeepMind}

\icmlcorrespondingauthor{}{alaas@deepmind.com, skapturowski@deepmind.com, dcalandriello@deepmind.com\vspace{-2\baselineskip}}

\icmlkeywords{Deep RL, exploration, density estimation, representation learning}

\vskip 0.3in
]

\printAffiliationsAndNotice{\icmlEqualContribution} 

\begin{abstract}
We introduce \textit{Robust Exploration via Clustering-based Online Density Estimation} (\DETOCS), a non-parametric method for novelty-based exploration that estimates visitation counts for clusters of states based on their similarity in a chosen embedding space.
By adapting classical clustering to the nonstationary setting of Deep RL, \DETOCS can efficiently track state visitation counts over thousands of episodes. We further propose a novel generalization of the inverse dynamics loss, which leverages masked transformer architectures for multi-step prediction; which in conjunction with \DETOCS achieves a new state-of-the-art in a suite of challenging 3D-exploration tasks in \DMH. 
\DETOCS also sets new state-of-the-art in hard exploration Atari games, and is the first agent to reach the end screen in \textit{Pitfall!}
\end{abstract}

\vspace{-\baselineskip}
\section{Introduction}

Exploration mechanisms are a key component of reinforcement learning (RL,~\citealp{sutton2018reinforcement}) agents, especially in sparse-reward tasks where long sequences of actions need to be executed before collecting a reward.
The exploration problem has been studied theoretically~\citep{kearns2002near,azar2017minimax,brafman2003r-max,auer2002finite, agrawal2012analysis,audibert2010best,jin2020reward} in the context of bandits~\citep{lattimore2020bandit} and Markov Decision Processes (MDPs,~\citealp{puterman1990markov,jaksch2010near}).
One simple yet theoretically-sound approach for efficient exploration in MDPs is to use a decreasing function of the visitation counts as an exploration bonus~\citep{strehl2008analysis,azar2017minimax}.
However, this approach becomes intractable for large or continuous state spaces, where the agent is unlikely to visit the exact same state multiple times, and some form of meaningful generalization over states is necessary.
Several approximations and proxies for visitation counts and densities have been proposed to make this form of exploration applicable to complex environments.
Two partially successful approaches in deep RL are: the parametric approach, which uses neural networks to estimate visitation densities directly, and the non-parametric approach, which leverages a memory of visited states to guide exploration.

\begin{figure}[t!]
    \centering
    \begin{subfigure}{.5\textwidth}
        \centering
        \includegraphics[width=0.9\textwidth]{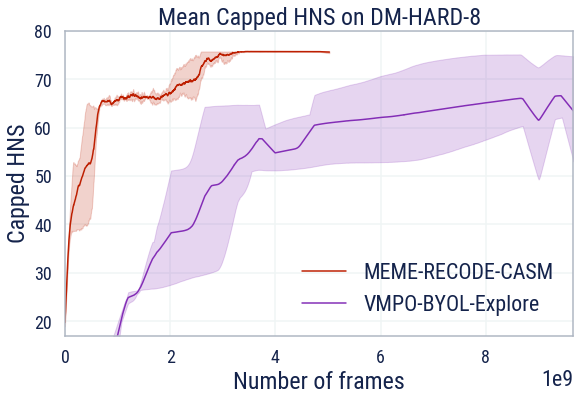}
    \end{subfigure}  

    \caption{
    A key result of \DETOCS is that it allows us to leverage more powerful state representations for long-term novelty estimation; enabling to achieve a new state-of-the-art in the challenging 3D task suite \DMH.
    \label{fig:teaser}\vspace{-1\baselineskip}
    }
\end{figure}

Parametric methods either explicitly estimate the visitation counts using density models~\citep{bellemare2016unifying,ostrovski2017count} or use proxies for visitation such as the prediction error of a dynamics model~\citep{pathak2017curiosity,guo2022byol}, or from predicting features of the current observation, e.g., features given by a fixed randomly initialized neural network as in 
\RND~\citep{burda2019exploration}. While this family of methods provides strong baselines for exploration in many settings~\citep{burda2018large}, they are prone to common problems of deep learning in continual learning scenarios, especially slow adaptation and catastrophic forgetting.
Parametric models trained via gradient descent are unsuitable for rapid adaptation (e.g., within a single episode) because it requires updates to the state representation before the exploration bonus can catch up.
Additionally, catastrophic forgetting makes parametric methods susceptible to the so-called `detachment' problem in which the algorithm loses track of promising areas to explore \citep{ostrovski2017count}.
Non-parametric methods rely on a memory to store encountered states~\citep{savinov2018episodic,badia2020never}.
This facilitates responsiveness to the most recent experience as well as preserving memories without interference.
However, due to computational constraints, it is necessary to limit the memory size which, in turn, requires a selection or aggregation mechanism for states.

To obtain the best of both worlds, Never Give Up (\NGU, \citealp{badia2020never}) combines a short-term novelty signal based on an episodic memory and a long-term novelty signal based on \RND into a single intrinsic reward.
However, the need to estimate two different novelty signals simultaneously adds complexity and requires careful tuning.
Moreover, as pointed out by~\citet{pathak2017curiosity}, the final efficacy of any exploration algorithm strongly depends on the chosen state representation.
If the state encoding is susceptible to noise or uncontrollable features in the observations, it can lead to irrelevant novelty signals and prevent meaningful generalization over states.
As \NGU relies on \RND for representation, it also inherits its encoding deficiencies in the presence of noisy observations which limits the applicability of the method in stochastic or complex environments.

In this paper, we tackle these issues by decomposing the exploration problem into two disentangled sub-problems.
First, (i) \textbf{Representation Learning} with an embedding function that encodes a meaningful notion of state similarity while being robust to uncontrollable factors in the observations.
Second, (ii) \textbf{Count Estimation} that is able to provide a long term visitation-based exploration bonus while retaining responsiveness to the most recent experience.
Addressing (i), we extend the inverse dynamic model proposed by~\citet{pathak2017curiosity} by leveraging the power of masked sequence transformers~\citep{devlin2018bert} to build an encoder which can produce rich representations over longer trajectories while suppressing the encoding of uncontrollable features.
We refer to our representation as \CASM, for \textit{Coupled Action-State Masking}.
In order to deliver on (ii) we introduce a novel, non-parametric method called Robust Exploration via Clustering-based Online Density Estimation (\DETOCS).
In particular, \DETOCS estimates soft visitation counts in the embedding space by adapting density estimation and clustering techniques to an online RL setting.
Our approach tracks histories of interactions spanning thousands of episodes, significantly increasing memory capacity over prior art in non-parametric exploration methods which typically only store the most recent history like the current episode.
In the presence of noise, we show that it strictly improves over state-of-the-art exploration bonuses such as \NGU or \RND.
\DETOCS matches or exceeds state-of-the-art exploration results on Atari and is the first agent to reach the end-screen in \textit{Pitfall!}.
Beyond 2D, our method also performs well in 3D domains and in conjunction with \CASM, sets new state-of-the-art results in the challenging \DMH suite (Fig.~\ref{fig:teaser}).

 \section{Background}
\label{sec:background}
We consider a discrete-time interaction~\citep{mccallum1995instance, hutter2004universal, hutter2009feature, daswani2013q} between an agent and its environment. 
At each time step $t\in\mathbb{N}$ the agent receives an observation $o_t\in\mathcal{O}$, that partially captures the underlying state $s\in \mathcal{S}$ of the environment and generates an action $a_t\in\mathcal{A}$. 
We consider policies $\pi: \mathcal{O}\rightarrow \Delta_\mathcal{A}$, that map an observation to a probability distribution over actions.
Finally, an extrinsic reward function $r_e: \mathcal{S}\times\mathcal{A}\rightarrow \mathbb{R}$ maps an observation to a scalar feedback. 
This function can be combined with an intrinsic reward function $r_i$ to encourage the exploratory behavior which might not be induced from $r_e$ alone.

The observations provided to the agent at each time step $t$ are used to build a representation of the state via an embedding function $f_\theta : \mathcal{O} \rightarrow \mathcal{E}$, associating $o_t$ with a vector $e_t=f_\theta(o_t)$.
Typically, the embedding space $\mathcal{E}$ is the vector space $\mathbb{R}^D$ where $D\in\mathbb{N}^*$ is the embedding size. 
Common approaches to learn $f_\theta$ include using an auto-encoding loss on the observation $o_t$~\citep{burda2018large}, an inverse dynamics loss~\citep{pathak2017curiosity}, a multi-step prediction loss at the latent level~\citep{guo2020bootstrap,guo2022byol}, or other similar representation learning methods.
In particular,~\citet{pathak2017curiosity} and \citet{badia2020never} highlight the utility of the inverse-dynamics loss to filter out noisy or uncontrollable features, e.g., an on-screen death timer as in \textit{Pitfall!}.

A popular and principled approach to exploration in discrete settings is to provide an intrinsic reward inversely proportional to the visitation count~\citep{strehl2008analysis,azar2017minimax}.
However, in large or continuous spaces the same state may be rarely encountered twice. \citet{badia2020never} remedy this issue by introducing a slot-based memory $M$, which stores all past embeddings in the current episode, and replaces discrete counts with a sum of similarities between a queried embedding $e_t=f_\theta(o_t)$ and its k-nearest-neighbors $\text{Neigh}_k(e_t)$ under the kernel $\mathcal{K}$:
\begin{equation}
\label{eq:reward}
r_{t} \propto \frac{1}{\sqrt{N(f_\theta(o_t))}}
\approx \frac{1}{\sqrt{\sum_{m \in \text{Neigh}_k(e_t)}{\mathcal{K}(e_t, m)}}}.
\end{equation}
Since storing the full history of embeddings throughout training would require a prohibitive amount of space, this slot-based memory is typically relegated to short-term horizons only, and in \NGU it is reset at the end of every episode. As a consequence, slot-based memory must be combined with a separate mechanism capable of estimating long-term novelty; resulting in additional method complexity and trade-offs.
In the following, we present a simple and efficient slot-based memory which can effectively track novelty over thousands of episodes.

 \section{\DETOCS}
\label{sec:DETOCS}
In this section we present our method, Robust Exploration via Clustering-based Online Density Estimation (\DETOCS), to compute intrinsic rewards for exploration.
\DETOCS takes inspiration from the reward of \NGU~\citep{badia2020never}, but while \NGU stores individual embedded observations in $M$ and uses periodic resets to limit space complexity, \DETOCS controls its space complexity by aggregating similar observations in memory. 
This requires storing a separate counter associated with each element in the memory and new observations need not be directly added to the memory, but will typically be assigned to the nearest existing element whose counter is then incremented.
Since the counters are never reset and the merged observations have a better coverage of the embedding space, \DETOCS's memory is much longer-term than a simple slot-based approach yielding state-of-the-art performance in many hard-exploration environments.
It also simplifies the estimation of novelty to only one mechanism vs.~two as in \NGU.
Moreover, the \DETOCS architecture is highly flexible, allowing it to be easily combined with a variety of RL agents and most importantly different representation learning methods.
As we show in the experiments, methods that can better leverage priors from learned representations, such as \DETOCS, outperform those that need to estimate novelty directly on raw observations, like \RND (and in turn \NGU).
We now present our detailed implementation and an overview of our techniques in Algorithm~\ref{alg:recode}.

\begin{algorithm}[tb]
\caption{\DETOCS}
\label{alg:recode}
\scriptsize
\setstretch{1.2}

\begin{algorithmic}[1]
\STATE {\bfseries Input: }Embedding $e$, Memory $M=\{m_l\}_{l=1}^{|M|}$, atom visitation counts $\{c_l\}_{i=l}^{|M|}$, number of neighbors $k$, relative tolerance to decide if a candidate new atom is far $\kappa$, squared distance estimate $\dm^2$, $\dm^2$'s decay rate $\tau$, discount $\gamma$, insertion probability $\eta$, kernel function $\mathcal{K}$, intrinsic reward constant $\irc$
\STATE {\bfseries Output: }Updated memory $M=\{m_l\}_{l=1}^{|M|}$, updated atom visitation counts $\{c_l\}_{i=l}^{|M|}$, updated squared distance $\dm^2$, intrinsic reward $r$
\STATE Compute $N_{\mathcal{K}}(M, e) = \sum_{l=1}^{|M|} (1 + c_l)\,\mathcal{K}(m_l, e)$;
\STATE Compute intrinsic reward $r=\left(\sqrt{N_{\mathcal{K}}(M, e)} + \irc\right)^{-1}$\;
\STATE Find nearest $k$ atoms to the embedding $e$: $\text{Neigh}_k(e)=\{m_j\}_{j=1}^k$\;
\STATE Update $\dm$ estimate: $\dm^2\gets (1-\tau)\,\dm^2 + \frac{\tau}{k}\sum_{m\in \text{Neigh}_k(e)} \|m-e\|^2_2$\;
\STATE Discount all atom counts $c_l \gets \gamma\, c_l\, \quad \forall l\in\{1,\cdots, |M|\}$\;
\STATE Find nearest atom $m_{\star} =  \argmin_{m \in M, m \neq m_j} \|m - e \|_2$\;
\STATE Sample uniformly a real number in $[0,1]$: $u\sim U[0, 1]$\;
\IF{$\| m_\star - e \|_2^2 > \kappa\, \dm^2$ {\rm and} $u < \eta$}
\STATE Sample atom to remove $m_j$ with probability $P(j) \propto 1 / c_j^2$ \comment{// Remove under-populated cluster}
\STATE Find atom $m_{\dag}$ nearest to $m_j$: $m_\dag = \argmin_{m \in M, m \neq m_j} \|m - m_j \|_2$
\STATE Redistribute the count of removed atom: $c_\dag \gets c_j + c_\dag$
\STATE Insert $e$ at index $j$ with count $1$: $m_j\gets {e}\, , c_j \gets 1$ \comment{// Create a new cluster}
\ELSE
\STATE Update nearest atom position $m_\star \gets \frac{c_\star}{c_\star+1}m_\star + \frac{1}{c_\star+1}e $
\STATE Update nearest atom count $c_\star \gets c_\star + 1$
\ENDIF
\end{algorithmic}
\end{algorithm}

\paragraph{Approximating visitation counts.}
Our estimator is based on a finite slot-based container $M=\{m_j\}_{j=1}^{|M|}$, where $|M|$ is the memory size. We refer to $m_j\in\mathcal{E}$ as atoms since they need not correspond to a single embedding as in \citet{badia2020never,badia2020agent57}
We also store a separate count vector $c$ such that $c_i$ is an estimate of the visitation count of $m_i$.
In particular, $c_i$ does not only reflect the number of visits to $m_i$ but also captures any previous visit sufficiently close to it.

Given a new embedding $e$, we estimate its \emph{soft-visitation count} (Alg.~\ref{alg:recode}:L3-4)  as the weighted sum of all atoms close to $e$ in the memory, according to a similarity kernel:
\begin{equation}
    N_\mathcal{K}(M, e) = \sum\nolimits_l (1+c_l) \mathcal{K}(m_l,e;\dm). 
\end{equation}
In particular, we choose our kernel function as:
\begin{equation}
\label{eq:our-kernels}
    \mathcal{K}(m_l, e) = \frac{1}{1 + \frac{\| e- m_l  \|_2^2}{\epsilon \dm^2}} \, \mathds{1}_{\left\{\| e- m_l  \|_2^2 < \dm^2\right\}}\, ,
\end{equation}
where $\epsilon\in\mathbb{R}_+$ is a fixed parameter. 
This definition is similar to \citet{badia2020never}, but we replace their sum over $e$'s top-$k$ neighbors with a sum over all atoms within a $\dm$ distance from $e$.
This choice prevents a counter-intuitive behaviour that can occur when using the $k$-NN approach with counts. In particular, it is desirable that the soft-visitation count of a given embedding should increase after adding it to the memory. However, adding atoms to the memory can change the $k$-NN list. If an atom displaced from this list has a large count, this might actually \emph{reduce} nearby soft-visitation count estimates instead of increasing them. 
Conversely, our approach is not affected by this issue.

Finally, we return the intrinsic reward $r$ as in~\eqref{eq:reward}, but add a small constant $\irc$ to the denominator for numerical stability and normalize $r$ by a running estimate of its standard-deviation as in~\citealp{burda2019exploration}.

\paragraph{Building the memory.}
To construct our memory we rely on the same aggregation principle we leveraged to estimate soft-visitation counts. 
In particular, we will draw a parallel between our atoms $m_i$ and the centroids of a clustering of observations.
We take inspiration from classical clustering and density estimation approaches such as $k$-means or DP-means \cite{kulis2011revisiting}; and adapt them
to deal with the challenges posed by our large scale RL setting:  memory size is limited and cannot store all past data, observations arrive sequentially, their distribution is non-stationary, and even the representation used to embed them changes over time.
We now describe how \DETOCS tackles these problems.

At every step we must update the memory $M$ to reflect the impact of seeing $e$ on the soft-visitation counts, while keeping the size $|M|$ fixed. 
Intuitively, two possible ways come to mind: either replace an existing atom with the new embedding, or update the position and count of an existing atom to be closer to $e$.
Let $m_\star$ be the closest atom to $e$ in $M$.
We adopt the following rules (Alg.~\ref{alg:recode}:L8-18) to integrate new embeddings into the memory, which are closely related to the DP-means clustering algorithm~\cite{kulis2011revisiting}:

\vspace{-.5\baselineskip}
\begin{itemize}
    \item  If $e$ satisfies $||m_\star-e||^2<
    \kappa \dm^2$, where  $\dm$ is an adaptive threshold and $\kappa>0$ a fixed parameter, it is ``assigned'' to the cluster encoded by $m_\star$ and we update $m_\star$'s value according to the convex combination of the counts of the existing embedding and the new one:
\begin{equation}
    \label{eq:cluster-update}
    m_\star \longleftarrow \frac{c_\star}{c_\star+1} m_\star + \frac{1}{c_\star+1} e
\end{equation}
    Its weight $c_\star$ is also incremented by $1$;
    \item If there is no close-by atom, we randomly decide whether to create a new one by flipping a coin with probability $\eta$. 
    If the coin-flip succeeds, we introduce the new embedding as a new atom, and we also remove an existing atom using a procedure described in the next paragraph.
    If the coin-flip fails, we instead update $m_\star$ as in \eqref{eq:cluster-update}.
\end{itemize}

\vspace{-.5\baselineskip}
The random coin-flip is introduced to increase the stability of the clustering algorithm to noise. In particular, an embedding far away from the memory will be inserted only after it is seen on average $1/\eta$ times, making one-off outliers less of a problem.
At the same time, once a far away embedding is observed multiple times and becomes relevant for the soft-visitation counts, there is a high chance that it will be added to improve the coverage of the memory. 
But to keep memory size finite, an existing atom must be removed. 
We investigate three different strategies to select an atom $m_i$ for removal based on its cluster count $c_i$: 
(a) removing with probability $\propto \frac{1}{c_i^2}$; 
(b) removing with probability $\propto \frac{1}{c_i}$; 
(c) removing the atom with the smallest $c_i$.
An ablation study over removal strategies in App.~\ref{app:non-stationary} (\cref{fig:removal_strategy,fig:removal_rule}),  empirically shows that strategy (a) works best for the settings we consider, but also that results are generally quite robust to the specific choice.

Whenever an atom $i$ is removed, its count $c_i$ is redistributed to the count of its nearest neighbor in order to preserves the total count of the memory.
The update rule of \DETOCS can be also interpreted from the theoretical point of view as an approximate inference scheme in a latent probabilistic clustering model. 
We provide a more detailed connection of our update rule with the DP-means algorithm in App.~\ref{appendix:clustering_analysis}. 

\begin{figure*}[ht]
\centerline{
\includegraphics[width=.65\textwidth, trim={0 1cm 0 1.5cm}, clip]{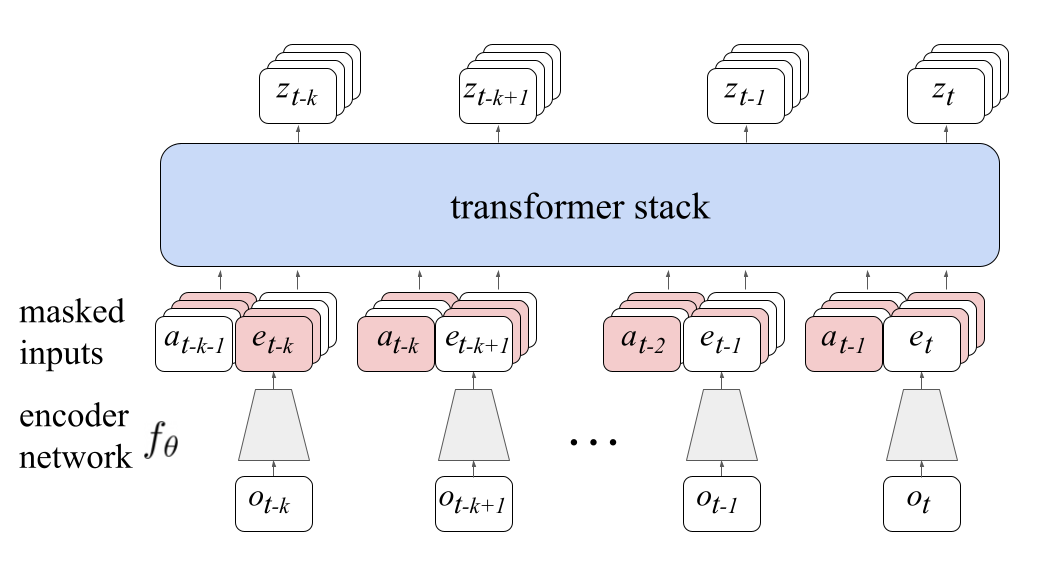}
\hfill
\includegraphics[width=.3\textwidth, trim={0 1cm 0 1.5cm}, clip]{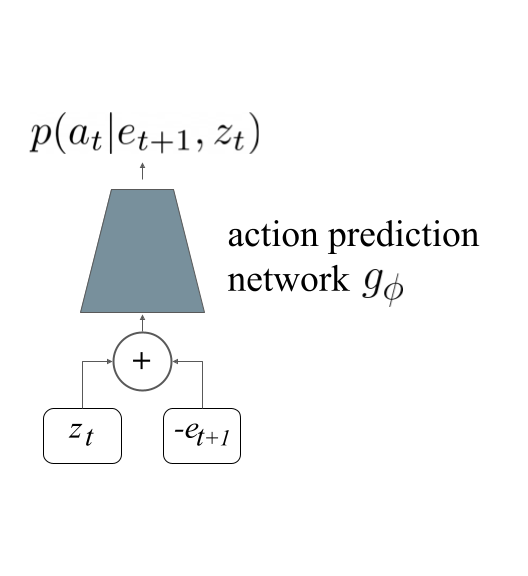}
}
\caption{Coupled Action-State Masking (\CASM) architecture used for learning representations in partially observable environments. 
The transformer takes masked sequences of length $k$ consisting of actions $a_i$ and embedded observations $e_i=f_\theta(o_i)$ as inputs and tries to reconstruct the missing embeddings in the output.
The reconstructed embeddings at time $t-1$ and $t$ are then used to build a 1-step action-prediction classifier.
The embedding function used as a representation for \DETOCS is $f_\theta$.
Masked inputs are shaded in pink, $N=4$ masked sequences are sampled during training (indicated by the stacks of $a$, $e$ and $z$ in the diagram). }\label{fig:architecture_representation_learning}
\end{figure*}

\paragraph{Dealing with non-stationary distributions.}
The distance scale between embedded observations can vary considerably between environments and throughout the course of training, as a result of non-stationarity in both the policy and embedding function $f_\theta$.
To deal with this issue, we include an adaptive bandwidth mechanism as in \NGU \cite{badia2020never}.
In particular, we update the kernel parameter $\dm^2$ whenever a new embedding $e$ is received,  based on the mean squared distance of the new embedding to the $k$-nearest existing atoms (Alg.~\ref{alg:recode}:L5-6). 
To allow for more rapid adaptation of $\dm$, we replace the running average used in \NGU with an exponential moving average with parameter $\tau$.

\begin{figure}[t]
    \centering
    \includegraphics[width=0.95\columnwidth]{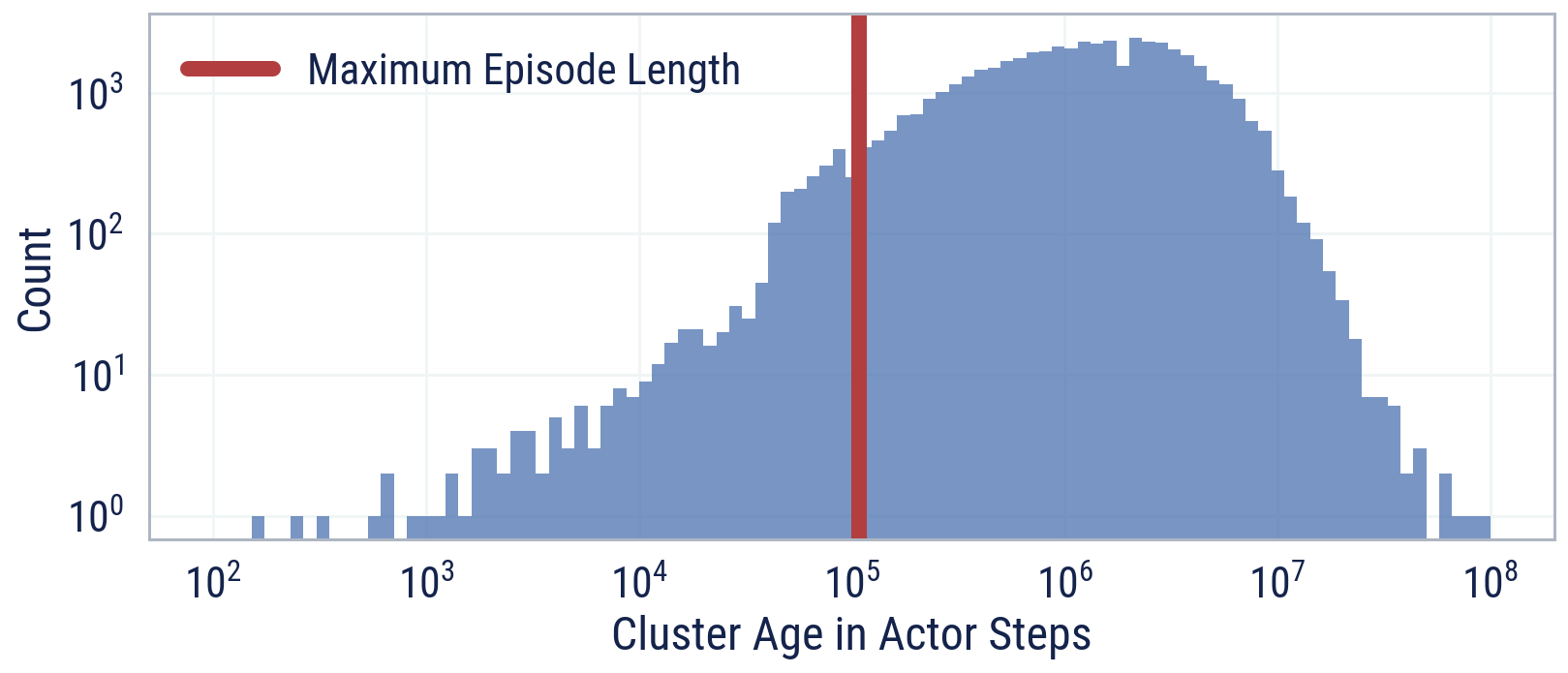}
    \caption{Content of an agent memory when it learns to reach \textit{Pitfall!}'s end screen.}
    \label{fig:pitfall_cluster_age_main}
\end{figure}

We note, however, that this mechanism is insufficient to cope with non-stationarity in $f_\theta$ over long timescales.
The original \NGU memory is not strongly impacted by this issue since it is reset after every episode, leaving little time for the representation to change significantly. 
However, in \DETOCS, these changing representations can end up corrupting the long-term memory if old clusters are not updated frequently.
In particular, an atom might achieve a high count under a representation, but become unreachable (and thus useless) under a different representation while still being unlikely to be removed. 
To counteract this we add a decay constant $\gamma$ which discounts the counts of all atoms in memory at each step as $c_i 
\longleftarrow \gamma c_i$, with $\gamma<1$ (Alg.~\ref{alg:recode}:L7).
This effectively decreases the counts of stale atoms over time and increases the likelihood of their removal during future insertions: clusters that do not get new observations `assigned' to them for a long time are eventually replaced.
At the same time, relevant clusters are kept alive much longer than previous methods.
Fig.~\ref{fig:pitfall_cluster_age_main} reports the histogram of cluster ages for clusters contained in the memory of an agent that has learned how to reach \textit{Pitfall!}'s end screen.
The red line in Fig.~\ref{fig:pitfall_cluster_age_main} denotes the maximum possible number of steps in an single episode, which is enforced by \textit{Pitfall!}'s in-game death timer, and would represent the maximum memory horizon for methods that reset their memory every episode.
As we can see, most of the clusters are much older than one episode, with earliest memories reaching back thousands of episodes.
We consider the effect of discounting in more detail in App.~\ref{app:non-stationary} (\cref{fig:devo_density_estimation_toy,fig:devo_cluster_ages,fig:pretrained_montezuma,fig:no_discount_atari}).

Importantly, we note that unlike \NGU where each actor maintains its own copy of the memory, \DETOCS shares the memory across all actors in a distributed agent, which greatly increases the frequency of updates to each atom resulting in less representation drift between memory updates.

 \begin{figure*}[ht]
    \centering
    \includegraphics[width=\textwidth]{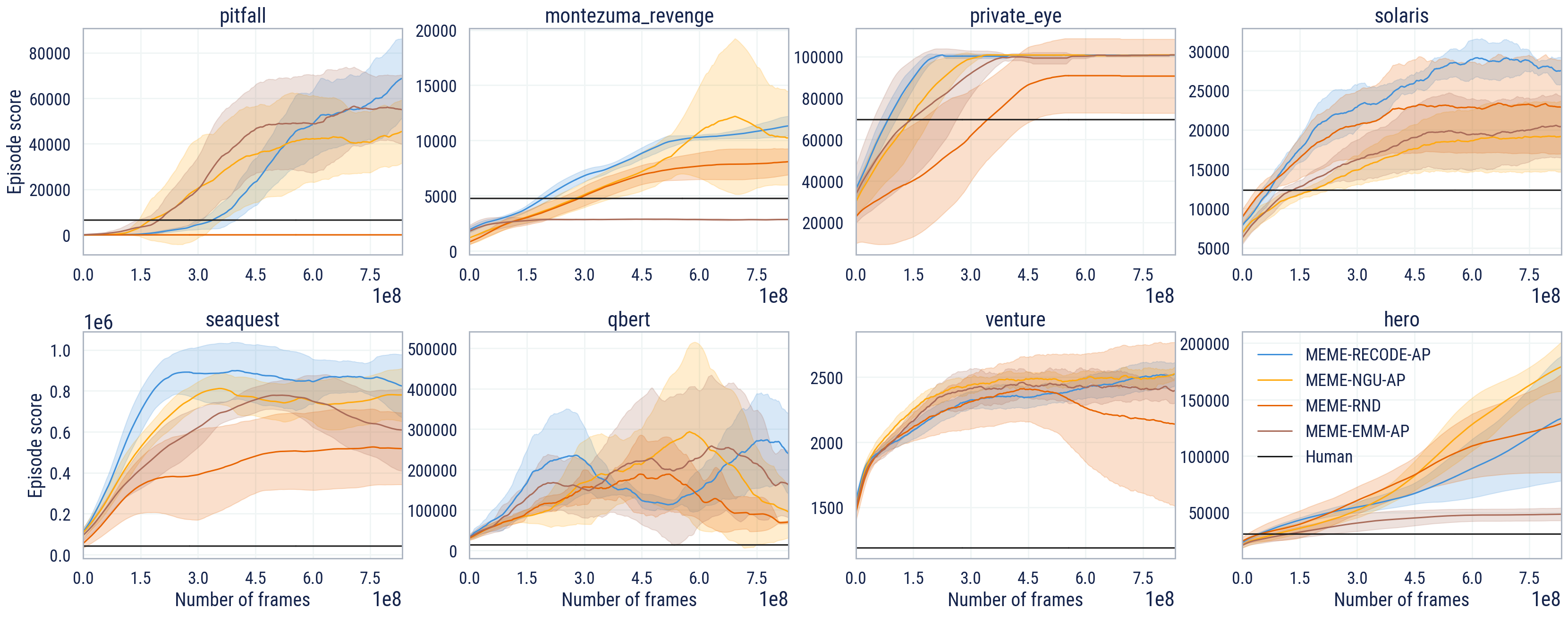}
    \caption{
    Comparison of \DETOCS against other exploration bonuses on Atari's hard exploration games.
    All agents are based on \MEME and use the same representation learning mechanism (\AP).
    Note that the high variance in \textit{Q*bert} is due to a bug in the game that, when exploited, allows to obtain significantly higher scores~\citep{chrabaszcz2018back}.}
    \label{fig:detocs_atari}
\end{figure*}

\section{Representation Learning Methods}
\label{subsec:representation_learning_methods}

As discussed in Section \ref{sec:background}, the choice of the embedding function $f_\theta: \mathcal{O} \rightarrow \mathcal{E}$ can have a significant impact on the quality of exploration; with many different representation learning techniques being studied in this context \citep{burda2018large,guo2020bootstrap,guo2022byol,guo2021geometric,erraqabi2021exploration}.
In the following, we focus on action prediction embeddings, introducing first the standard $1$-step prediction formulation~\citep{pathak2017curiosity,badia2020never,badia2020agent57}. 
The embedding function $f_\theta$ is parameterized as a feed-forward neural network taking $o_t$, the observation at time $t$, as input. 
We define a classifier $g_\phi$ that, given the embeddings of two consecutive observations $f_\theta(o_t), f_\theta(o_{t+1})$, outputs an estimate $p_{\theta, \phi}(a_t| o_t, o_{t+1}) = g_{\phi}\left( f_{\theta}(o_t), f_\theta(o_{t+1})\right)$ of the probability of taking an action given two consecutive observations $(o_t, o_{t+1})$. 
Both $f_\theta$ and $g_\phi$ are then jointly trained by minimizing an expectation of the negative log likelihood:
\begin{equation}
    \min_{\theta, \phi} \mathcal{L}(\theta, \phi)(a_t)=-\ln(p_{\theta, \phi}(a_t|o_{t}, o_{t+1}))\, ,
\end{equation}
where $a_t$ is the true action taken between $o_t$ and $o_{t+1}$. 
These embeddings proved to be helpful in environments with many uncontrollable features in the observation~\citep{badia2020never}, such as in Atari's \emph{Pitfall!}, where the observations contain many spurious sources of novelty even when the agent is standing still.

While \DETOCS can be used with an arbitrary embedding function, e.g. one tailored for the domain of interest, the choice of a meaningful representation is also a key factor for the final performance.
A major downside of the standard, $1$-step action-prediction method is the  simplicity of the prediction task, which can often be solved by learning highly localized and low-level features (e.g. how a single object shifts under a transition), which need not be informative of the global environment structure. In contrast, an ideal embedding should capture higher-level information about the environment, such as the agent's position or relative location of previously observed landmarks; which might not be simultaneously present in the individual observations $o_t$ and $o_{t+1}$. In order to achieve this, a wider context of time-steps may be needed.

However, the prediction task would become even easier if we simply provided the full trajectory to the predictor.
In order to address this limitation, we propose to use a \emph{stochastic} context, $h_t$, where at each timestep $k\leq t$, either $f_\theta(o_k)$ or $a_{k-1}$ is provided.\footnote{We avoid masking both $f_\theta(o_k)$ and $a_{k-1}$ simultaneously as this would increase the likelihood that the prediction task is indeterminable.}
The main intuition being that the model can still predict $a_t$ by learning to infer the missing information from $f_\theta(o_t)$ given $(h_{t-1}, a_{t-1})$. 
In this way, the action predictor would not solely rely on the information provided by $f_\theta(o_t)$, but it would also construct redundant representations within $h_t$.

From an implementation standpoint, we first build a sequence of observation embeddings and actions, $(f_\theta(o_0), a_0, f_\theta(o_1), \dots, a_{t-1}, f_\theta(o_t))$. Then, inspired by masked language models~\citep{devlin2018bert}, at each timestep $t$, we randomly substitute either $f_\theta(o_t)$ or $a_t$ with a special token indicating missing information. 
These masked sequences are then fed to a causally-masked transformer, whose output is then projected down to the size of the embedding ($\dim z_t = \dim f_{\theta}(o_t)$), and the difference between the two is input into a final MLP classifier $ g_{\phi}$.
As with $1$-step action prediction, we train the representation using maximum likelihood. We refer to this approach as Coupled Action-State Masking (\CASM) in the following.
During training, we randomly sample multiple masked sequences per trajectory ($N=4$) to help reduce gradient variance. 
Note that the final embedding that we provide to \DETOCS is $e_t=f_{\theta}(o_t)$, i.e. the transformer \textit{inputs}, to avoid leaking information about the agent's trajectory.
Figure~\ref{fig:architecture_representation_learning} shows a diagram of the architecture.
 \begin{figure*}[ht]
    \centering
    \includegraphics[width=0.99\textwidth]{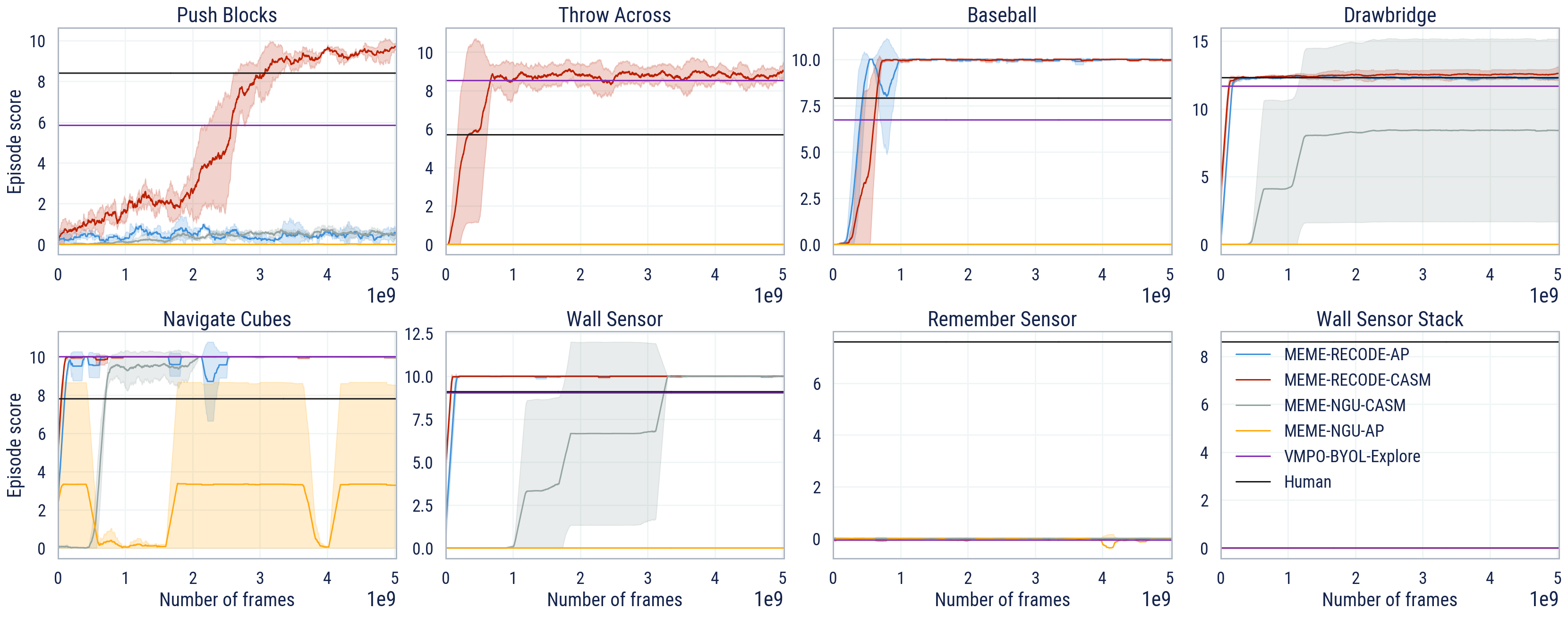}
    \caption{
    Performance of \DETOCS compared to \NGU and \BYOLE on the single-task version of \DMH.
    The BYOL-Explore results correspond to the final performance reported in~\citet{guo2022byol} after \texttt{1e10} environment frames.
    All results have been averaged over 3 seeds. \label{fig:dmh8_single_task}
    }
    \vspace{-.2cm}
\end{figure*}

\section{Experiments}
\label{sec:experiments}

In this section, we experimentally validate the efficacy of our approach on two established benchmarks for exploration in 2D and 3D respectively: a subset of the Atari Learning Environment (ALE, \citealp{bellemare2013arcade}) containing eight games such as \texttt{Pitfall} and \texttt{Montezuma's Revenge} which are considered hard exploration problems~\citep{bellemare2016unifying}; and \DMH~\citep{gulcehre2019making}, a suite of partially observable 3D games.
All games pose significant exploration challenges such as very long horizons ($\mathcal{O}$(10K) steps), the necessity to backtrack, sparse rewards, object interaction and procedural environment generation.
Our method achieves state-of-the-art results across both benchmarks and even solves two previously unsolved games: in Atari's \textit{Pitfall!} our method is the first to reach the end screen and on \DMH's \textit{Push Block} we are the first to achieve super-human performance.
We also perform a set of ablations to shed more light on the influence of the representation learning mechanism and the robustness w.r.t. noisy observations.

All candidate architectures evaluated in the following experiments (and in App.~\ref{appendix:ablations}), are composed of three main modules:
(1) a base agent, responsible for core RL tasks such as collecting observations and updating the policy,
(2) an algorithm responsible for generating the exploration bonus, and
(3) an embedding mechanism responsible for learning meaningful representations of observations.
Our nomenclature reflects the choice of modules as \texttt{AGENT-EXPLORATION-EMBEDDING}.
For example, the \MEME agent described in~\citet{kapturowski2022human} is denoted as \MEME-\NGU-\AP.
We use the \MEME agent across all experiments, but vary the exploration and representation mechanisms.
For exploration we consider \EMM (pure episodic memory), \NGU and \DETOCS whereas for representation we experiment with \AP and \CASM.
We provide a full list of hyper-parameters for all agents and baselines in App.~\ref{appendix:hypers}.

\subsection{Atari}
\label{subsec:atari_experiments}

The hard-exploration subset of Atari as identified by~\citet{bellemare2016unifying} poses a considerable challenge in terms of optimization horizon with episodes lasting up to $27,000$ steps using the standard action-repeat of four.
Additionally, rewards vary considerably in both scale and density.
Across all our experiments in the Atari domain, we set the memory size of our agent to $5\cdot 10^4$ atoms.
We evaluate all agents following the regime established in prior work~\citep{mnih2015human,van2016deep} using $30$ random no-ops, no `sticky actions'~\citep{machado2018revisiting} and average performance over six seeds.

We compare the game scores obtained using our exploration bonus, \DETOCS, against other methods while keeping agent architecture and representation mechanism fixed.
The results presented in Fig.~\ref{fig:detocs_atari} show that our method achieves state-of-the-art, super-human performance across all eight
games while using a conceptually simpler exploration bonus compared to \MEME-\NGU-\AP.
The \MEME-\EMM-\AP and \MEME-\RND ablations in Fig.~\ref{fig:detocs_atari} reveal the respective shortcomings of short-term and long-term novelty when used in standalone fashion.
\EMM on its own cannot solve \textit{Montezuma's Revenge} because it requires long-term memory.
Conversely, \RND on its own cannot solve \textit{Pitfall!} because of the presence of many uncontrollable features in the observations and its inability to leverage the \AP embeddings.
In contrast, \DETOCS is able to leverage the \AP representation for short-term and long-term novelty due to the clustering-based memory integrating over a long horizon which enables solving both games with a single intrinsic reward.

\subsection{\DMH}
\DMH~\citep{gulcehre2019making} consist of eight exploration tasks, designed to challenge an RL agent in procedurally-generated 3D worlds with partial observability, continuous control, sparse rewards, and highly variable initial conditions.
Each task requires the agent to interact with specific objects in its environment in order to reach a large apple that provides reward (cf. Fig.~\ref{fig:agi_human_knock_obj_demo} in the Appendix for an example).
The procedural generation randomizes object shapes, colors, and positions at every episode.
Across all our experiments in the \DMH domain, we set the memory size of our agent to $2\cdot 10^5$ atoms.
We also use the more powerful \CASM representation over \AP as the default in these experiments but present an ablation on the influence of the representation in Sec.~\ref{sec:ablations}.
All performances reported for evaluation are averaged across three seeds.

We compare \DETOCS with \NGU and the recently proposed \BYOLE~\citep{guo2022byol} in this domain.
The results presented in Fig.~\ref{fig:dmh8_single_task} show that our method is able to solve six out of eight tasks with super-human performance which sets a new state-of-the-art on this benchmark and marks the first time that the human baseline has been beaten on \textit{Push Blocks}.
To control for the contribution of the representation, we also run a version of \NGU which uses the more powerful \CASM representation instead of its default \AP one.
Switching \AP with \CASM improves \NGU's performance significantly and stresses the importance of incorporating information over longer trajectories in the representation mechanism for this domain to combat the challenge of partial observability.
However, only \DETOCS is able to take full advantage of the representational power afforded by \CASM as it is able to leverage it for both short-term and long-term novelty bonuses.

\subsection{Ablations}
\label{sec:ablations}
Concluding our experiments, we perform two ablation studies to gauge the sensitivity of our approach to the presence of noisy observations and the choice of the underlying representation mechanism.

\begin{figure}[t!]
\centering
\begin{subfigure}{.5\textwidth}
    \centering
    \includegraphics[width=0.75\textwidth]{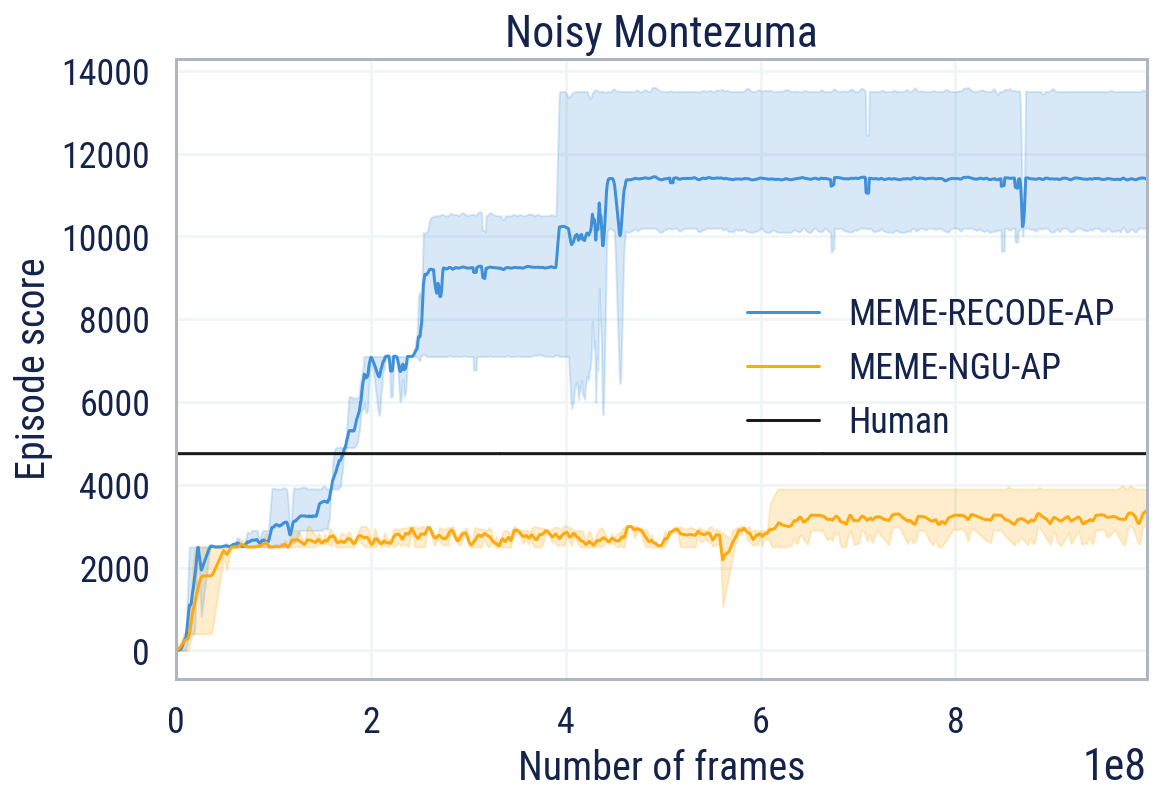}
\end{subfigure}  
\begin{subfigure}{.2\textwidth}
    \centering
    \includegraphics[width=0.9\textwidth]{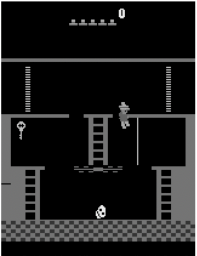}
\end{subfigure}
\begin{subfigure}{.2\textwidth}
    \centering
    \includegraphics[width=0.9\textwidth]{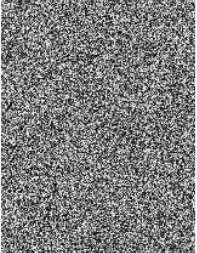}
\end{subfigure}  
\caption{Robustness to observation noise.
Top: Performance of \DETOCS compared to \NGU on Noisy Montezuma.
Bottom: A frame of Noisy Montezuma where the noise is concatenated to the original frame.
\label{fig:detocs_noisy_montezuma_performance}}
\vspace{-.2cm}
\end{figure}

\vspace{-.2cm}
\paragraph{Robustness to observation noise.}
Noise in the observation space is one of the most significant adversarial conditions exploration methods must to overcome to deliver utility for any practical scenario which always features imperfect sensors.
The `noisy TV problem'~\citep{schmidhuber2010formal,pathak2017curiosity} is a common metaphor which describes a failure mode of exploration methods getting stuck on the prediction of noise as a meaningless signal of novelty.
In order to assess our method's robustness w.r.t. observation noise, we construct a noisy version of \textit{Montezuma's Revenge} by concatenating a frame containing white noise in the range $[0, 255]$ to the game's original $210\times 160$ greyscale observations along the image height dimension.
We compare \DETOCS to \NGU in this setting using the same \AP backbone to suppress uncontrollable noise on the representation level and assess the sensitivity of the exploration bonus to it.
The results of this experiment are presented in Fig.~\ref{fig:detocs_noisy_montezuma_performance}.
We find that the performance of \MEME-\NGU-\AP deteriorates significantly in the presence of noise.
This can be attributed to the fact that \NGU relies on \RND to compute the long-term exploration bonus, which degenerates to random exploration in the presence of uncontrollable noise~\citep{kapturowski2018recurrent}.
This effectively restricts the baseline to short-term exploration within one episode.
In contrast, \DETOCS's mean performance is not degraded significantly and achieves a similar score as in Fig.~\ref{fig:detocs_atari}, albeit with a higher variance.

\begin{figure}
    \centering
    \includegraphics[width=0.42\textwidth]{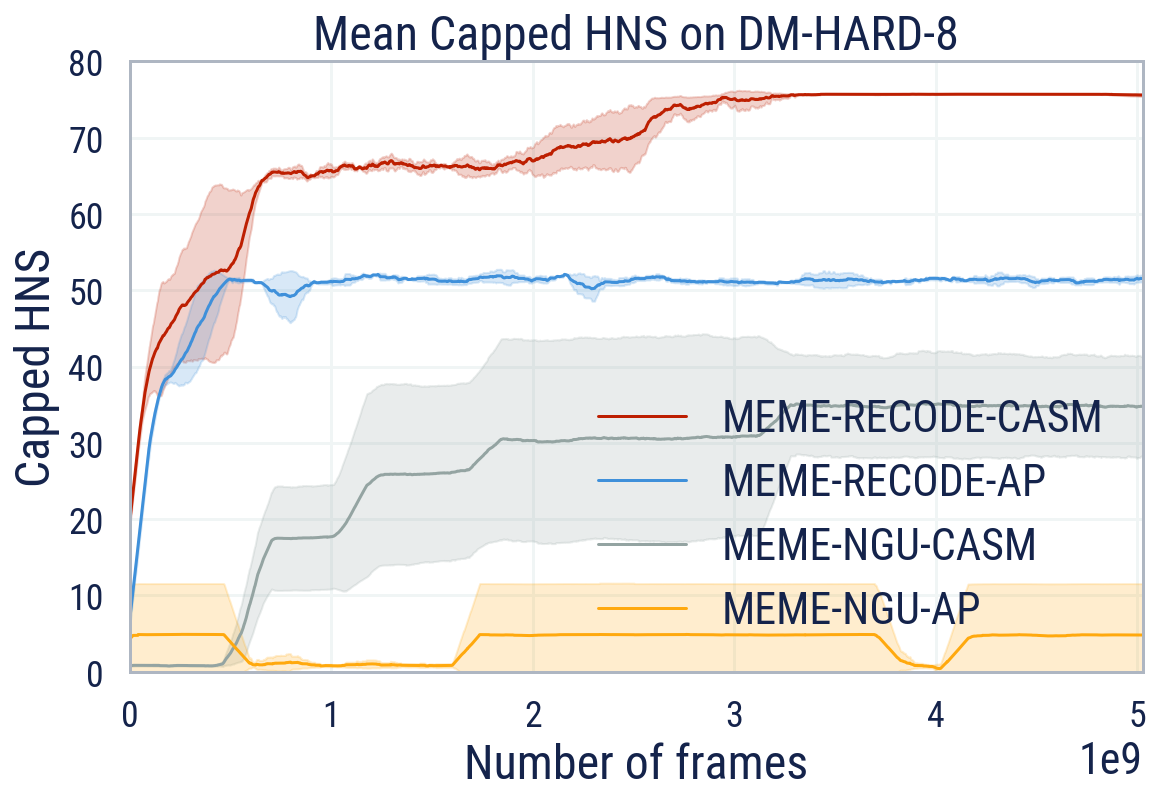}
    \caption{
    Comparing \DETOCS and \NGU leveraging \AP and \CASM representations across all \DMH tasks using capped human-normalized scores. 
    \label{fig:dmh8_representation_ablation}
    \vspace{-.25cm}
    }
\end{figure}

\paragraph{Leveraging different representation mechanisms.}
The experiments on \DMH demonstrate the importance of employing more powerful representation learning techniques in more complex, partially observable environments.
However, while a richer representation often provides a flat boost to downstream task learning, it cannot solve the exploration problem in itself.
In Fig.~\ref{fig:dmh8_representation_ablation}, we compare the contribution of \AP and \CASM to the aggregated performance of \NGU and \DETOCS on \DMH.
The results consistently demonstrate that \CASM is a superior representation to \AP in this domain, leading to significant performance gains with both exploration methods.
However, \DETOCS outperforms \NGU for both representations, indicating that leveraging the representational power for both short-term and long-term novelty signals is a key benefit of our proposed method.
 \section{Conclusion}
In this paper, we introduce Robust Exploration via Clustering-based Online Density Estimation (\DETOCS), a principled yet conceptually simple exploration bonus for deep RL agents that allows to perform robust exploration by estimating visitation counts from a slot-based memory.
\DETOCS improves over prior non-parametric exploration methods by increasing the effective memory span by several orders of magnitude using an online clustering mechanism.
Our method sets a new state-of-the-art in task performance on two established exploration benchmarks, Atari's hard exploration subset and \DMH.
It is also the first agent to reach the end screen in \textit{Pitfall!} within the time limit which exemplifies \DETOCS's efficiency of leveraging both long-term (i.e. previous experience) and short-term (i.e. within an episode) novelty signals.
Beyond the benchmarks, \DETOCS's performance also remains unaffected by noisy observations -- an adversarial condition which significantly degrades prior approaches such as \RND and \NGU.
Additionally, we show that our method is agnostic to the concrete representation technique chosen for embedding the observations and scales well with increasingly powerful representations, e.g. using multi-step sequence prediction transformers like our proposed \CASM architecture.
We conclude that \DETOCS can serve as a simple yet robust drop-in exploration method compatible with any RL agent and representation learning method which directly translates improvements in representation learning to improvements in exploration performance.

\bibliographystyle{icml2023}

\newpage
\appendix
\onecolumn
\section{Related Works}
\label{appendix:related_works}
In this section, we give a brief and non-exhaustive overview of past works computing visitation counts or estimating densities in RL. We classify them as either parametric or non-parametric. 

\paragraph{Parametric methods.}~\citet{bellemare2016unifying} and \citet{ostrovski2017count} propose to compute pseudo-visitation counts using density estimators on images such as Context Tree Switching (CTS,\citealp{bellemare2014skip}) or PixelCNN~\citep{van2016conditional}. On the other hand, \citet{tang2017count} use locality-sensitive hashing to map continuous states to discrete embeddings, where explicit visitation counts are computed. Some methods such as \RND~\citep{burda2019exploration} can be interpreted as estimating implicitly the density of observations by training a neural network to predict the output of a randomly initialized and untrained neural network which operates on the observations. \citet{hazan2019provably, pong2019skew, lee2019efficient, guo2021geometric} propose algorithms that search a policy maximizing the entropy of its induced state-space distribution. In particular, the loss optimized by \citet{guo2021geometric} allows to compute a density estimate as well as maximizing the entropy. Finally, \citet{domingues2021density-based} computes a density estimation on top of learned representations, which are inspired by bonuses used in reward-free finite MDPs.

\paragraph{Non-parametric methods.}
Non-parametric density estimates that we build on date back to~\citet{Rosenblatt1956,Parzen1962} 
(Parzen–Rosenblatt window) and are widely used in machine learning as they place very mild assumptions on the data distribution. Non-parametric, kernel-based approaches have been already used in RL and shown to be empirically successful on  smaller-scale environments by \citet{kveton2012kernel} and \citet{barreto2016practical} and are theoretically analyzed by~\citet{ormoneit2002kernel, pazis2013pac, domingues2020regret}. In \NGU~\citep{badia2020never}, Agent57~\citep{badia2020agent57} and \MEME~\citep{kapturowski2022human}, a non-parametric approach is used to compute a short term reward at the episodic level.
\citet{liu2021behavior} propose an unsupervised pre-training method for reinforcement learning which explores the environment by maximizing a non-parametric
entropy computed in an abstract representation space. The authors show improved performance on transfer in Atari games and continous control tasks. \citet{seo2021state} use random embeddings and a non-parametric approach to estimate the state-visitation entropy, but do not generalize to concurrently learned embeddings.
\citet{tao2020novelty} show that K-NN based exploration can improve exploration and data efficiency in model-based RL.
While non-parametric methods are good models for complex data, they come with the challenge of storing and computing densities on the entire data set. We tackle this challenge in Sec.~\ref{sec:DETOCS} of the main text by proposing a method that estimates visitation counts over a long history of states, allowing our approach to scale to much larger problems than those considered in previous works, and without placing assumptions on the representation, that can be trained concurrently with the exploration process and doesn't need to be fixed a priori. \section{General notation}
\label{appendix:notation}
We consider the usual Reinforcement Learning setting, where an agent interacts with an environment to maximize the sum of discounted rewards, with discount $\gamma\in[0,1)$, as in \citealp{sutton1988}.
In particular, the environment can be described as a Partially-Observable Markov Decision Process (POMDP) \cite{kaelbling1998planning}. 
First, we define a Markov Decision Process (MDP) through a tuple $(\mathcal{S},\mathcal{A},T,R)$, where $\mathcal{S}$ is the set of states, $\mathcal{A}$ is the set of possible actions, $T$ a transition function, which maps state-actions to distributions over next states, and $R:\mathcal{S}\times \mathcal{A}\to \mathbb{R}$ is the reward function.
In particular, a Markov Decision Process is a discrete-time interaction process \cite{mccallum1995instance,hutter2004universal, hutter2009feature, daswani2013q} between an agent and its environment.
In a Partially-Observable MDP, the agent does not receive a state from $\mathcal{S}$, but an observation $o\in \Omega$, where $\mathcal{O}$ is the function mapping unobserved states to distributions over observations. An observation $o$ will only contain partial observations about the underlying state $s\in\mathcal S$. 
This function can be combined with an intrinsic reward function $r_i$ to enable the exploratory behavior.
The environment responds to an agent's action $a\in\mathcal{A}$ by performing a transition to a state $s'\sim T(\cdot | s,a)$; the agent receives a new observation $o'\sim\Omega(\cdot|s')$ and a reward $r\sim R(s,a)$.
At step $t$, we can indicate with $h_t = \{o_0, a_1, o_1, \ldots, a_t, o_t\} \in \mathcal{H}_t$ the history of past observations-actions, where
$\mathcal{H}_{t}=\mathcal{H}_{t-1}\times\mathcal{A}\times\mathcal{O}$, $\mathcal{H}_0=\mathcal{O}$
and the overall history space is $\mathcal{H}=\bigcup_{t\in\mathbb{N}}\mathcal{H}_t$.
We consider policies $\pi: \mathcal{H}\rightarrow \Delta_\mathcal{A}$, that map a history of past observations-actions to a probability distribution over actions. 

 \section{\DETOCS from a Clustering Point of View}
\label{appendix:clustering_analysis}

The update rules \DETOCS's memory structure in Algorithm~\ref{alg:recode} of the main text can be interpreted as an approximate inference scheme in a latent probabilistic clustering model.
We explore this connection here as means to better understand and justify the proposed algorithm as a density estimator.
The rule has a close connection to the DP-means algorithm of \cite{kulis2011revisiting}, with two key differences:
\begin{itemize}
\itemsep0em 
    \item the counts of the cluster-centers are discounted at each step, allowing our approach to deal with the non-stationarity of the data due to changes in the policy and the embedding function, effectively reducing the weight of stale cluster-centers in the memory,
    \item when creating a new cluster-center, we remove an underpopulated one, so as to keep the size of the memory constant. 
\end{itemize}

The adaptations are necessary to accommodate the additional complexities of our setting, which follows a streaming protocol (i.e. data must be explicitly consumed or stored as it arrives, and data that are not stored cannot be accessed again) and is non-stationary (i.e. data are not assumed to be identically distributed as time advances). 
The clustering algorithm resulting from these adaptations is shown in Algorithm~\ref{alg:clustering}. \DETOCS implements such an algorithm to update the memory, and it also calculates an intrinsic reward for the observed embedding $e$, as described in Section \ref{sec:DETOCS} .

\begin{algorithm}[tb]
\caption{A streaming clustering algorithm.}
\label{alg:clustering}
\setstretch{1.2}
\begin{algorithmic}[1]
\STATE {\bfseries Parameters: }\\
Number of clusters $|M|$ \\
Number of nearest cluster centres $k$ \\
Discounting of counts at each step $\gamma$\\
Distance threshold to propose the creation of a new cluster  $\kappa$ \\
Probability of accepting the creation of a new cluster $\eta$\\

\STATE {\bfseries State: }\\
Threshold to create new cluster (i.e. average cluster distance) $d = 0$ \\
Cluster centres $m_l = 0\quad \forall{l \in 1\dots |M|}$\\
Cluster counts $c_l = 0\quad \forall{l \in 1\dots |M|}$\\
Indices of $k$-nearest neighbours of point $e$: $\text{Neigh}_k(e)$ \\

\STATE {\bfseries Implementation: }\\
\FORALL{received embedding $e \in \{e_0, e_1, e_2, \dots \}$}
\STATE Update average inter-cluster distance $d \gets (1-\tau) d + \frac{\tau}{k}\sum_{l \in \text{Neigh}_k(e)} \|m_l - e\|^2_2$ 
\STATE Discount all cluster-center counts $c_l \gets \gamma\, c_l\, \quad \forall l\in 1,\dots, |M|$
\STATE Find index of nearest cluster center $m_\star =  \argmin_{m\in M} \|m_l - e \|_2$
\IF{$\| m_i - e \|_2^2 > \kappa\, d$ \text{and with probability $\eta$}}
\STATE Sample index $j$ of cluster center to remove with probability $P(j) \propto 1 / c_j^2$
\STATE Find index of nearest cluster center to $m_j$: $m_\dag = \argmin_{m\in M, l\neq j} \|m_l - m_j \|_2$
\STATE Redistribute the counts of removed cluster center: $c_\dag \gets c_j + c_\dag$
\STATE Replace cluster $j$ with a the new cluster at $e$: $m_j\gets {e}\, , c_j \gets 1$
\ELSE 
\STATE Update nearest cluster center $m_\star \gets \frac{c_\star}{c_\star+1}\mu_i + \frac{1}{c_\star+1}e $
\STATE Update nearest cluster-center count $c_\star \gets c_\star + 1$
\ENDIF

\ENDFOR
\end{algorithmic}
\end{algorithm}

\subsection{Addressing finite-memory limitations. }
\label{app:finite-memory}
We first address the modifications introduced to deal with the memory limitations of the streaming setting: 1) each datum (embedding $e_t$ in our notation) is incorporated into a cluster distribution approximation once, then discarded; 2) the total number of clusters is stochastically projected down onto an upper limit on the number of clusters (otherwise they would grow without bound--albeit progressively more slowly).
Both modifications allow our method to maintain constant space complexity in the face of an infinite stream of data.

The \emph{step-wise} justification of the Algorithm~\ref{alg:clustering} is relative straightforward. At step $t$, for embedding $e_t$, we show that the following objective is minimised:
\begin{align}
\label{eq:stepwise}
    &\min_{l\in 1, \dots, |M|} \|m_l - e_t\|^2_2 \\ 
\nonumber
    &\text{s.t.}\quad \|m_l - e_t\|^2_2 \le \kappa d
\end{align}
Working backwards: updating the cluster center  reduces the objective directly and will not violate the constraint (unless it was already in violation; this excluded in the precondition of this branch). This accounts for the ``else'' branch.
The ``if'' branch introduces a new cluster center precisely at $e_t$, thus \eqref{eq:stepwise} is minimised completely: it is zero for this branch.
Finally, selecting the index of the nearest cluster center directly minimises placement of the branch according to \eqref{eq:stepwise}, ignoring the constraint (which is latest ensured by the ``if/else'').
Note that the hard constraint of \eqref{eq:stepwise} takes the place of the soft cluster penalty of DP-means \citep{kulis2011revisiting}.

The updates to the cluster centers, unlike k-means and DP-means, are done in an exponentially-weighted moving average of the embeddings, rather than as global optimisation step utilising all of the data.
Consequently, and importantly, what happens to \eqref{eq:stepwise} evaluated for $e_s$, where $s\ne t$, is of significant interest, as objectives for k-means and DP-means account for all data, rather than a single datum.

\begin{figure}[H]
\centering
\includegraphics[width=0.8\textwidth]{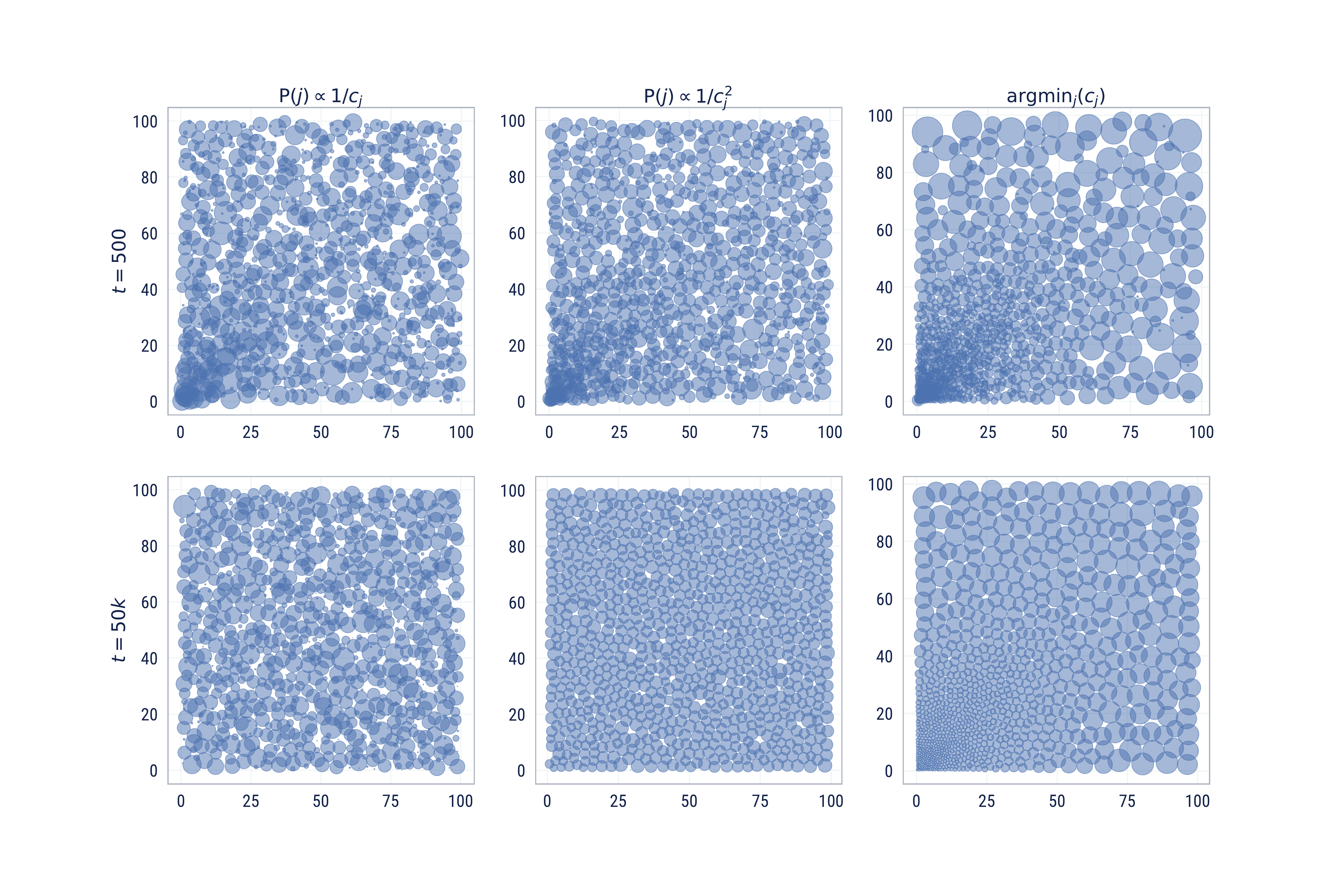}
\caption{Effect of removal strategy on evolution of cluster centers and counts (with counts corresponding to the size of the marker).
At each timestep $t$ we sample a batch of 64 2D-embeddings from a square of side $\min(100, t)$. 
After $t=100$ the distribution remains stationary and we would like the distribution of cluster centers and counts to be to become approximately uniform after enough time has passed. For a deterministic removal strategy which selects the clusters with the lowest counts, the cluster centers can remain skewed long after the distribution has stopped changing. 
For both probabilistic removal strategies, the cluster centers become approximately uniform, but only for the $1/c_j^2$ removal strategy we observe that both cluster centers and counts become uniform. Note that we use a discount of $\gamma = 0.9999$.}  \label{fig:removal_strategy}
\end{figure}

\begin{figure}[H]
\centerline{
\includegraphics[width=.9\textwidth]{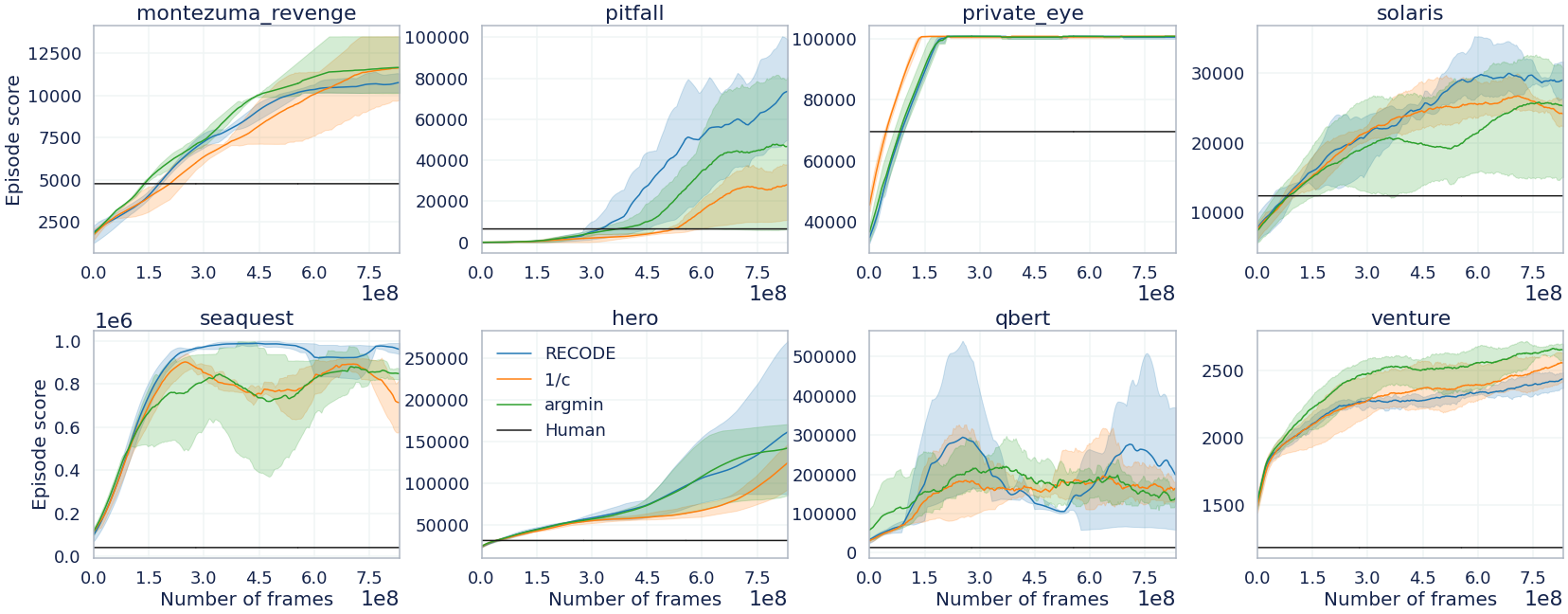}
}
\caption{Effect of removal strategy on performance. All choices of removal strategy considered result in a viable algorithm, but there are some environments (most notably \emph{Pitfall!}) where the chosen strategy of $1/c^2$ appears to be more robust.}\label{fig:removal_rule}
\end{figure}

We tested the qualitative behavior of different removal strategies in Fig.~\ref{fig:removal_strategy}. This study suggested that a stochastic removal of a cluster with probability $\propto c^{-2}$ was more stable and better tracked a non-stationary distribution.
The intuition we got from these toy examples is also confirmed in ablation experiments ran on the Atari environment, as shown in Fig.~\ref{fig:removal_rule}, where we compare \DETOCS runs with different removal rules.

\subsection{Dealing with non-stationary embedding distributions. }
\label{app:non-stationary}
We now turn to the question of how to deal with the non-stationarity of the embedding distribution. 
We introduced the following modifications to deal with the non-stationarity of our data stream: 
\begin{enumerate}
\itemsep0em 
    \item the cluster count decays,
    \item two clusters can be merged to accommodate a new one,
    \item the use of an exponentially weighted moving average update of cluster centers.
\end{enumerate}

In k-means, all of the data are retained. 
This makes k-means costly: at each step of fitting the entire data set is examined to update the cluster assignments and update the cluster means.
Instead, we take a distributional approximation to the data associated with each cluster, and when re-adjusting cluster assignments according to \eqref{eq:stepwise}, we do so in terms of this distributional approximation.

In particular, each cluster is approximated by a Gaussian distribution with precision 1 and whose mean is unknown but with prior zero and precision 1. Specifically:
\begin{align*}
    \mu_l &\sim \mathcal{N}(0, c_0) \\
    e_i|\mu_l &\sim \mathcal{N}(\mu_l, 1)
\end{align*}
where $\mathcal{N}(\mu, \tau)$ denotes a Gaussian (or normal) distribution with mean $\mu$ and precision $\tau$ (precision is the inverse variance).
Since the prior on $\mu_l$ is conjugate to the likelihood on $e_i$, we know that the posterior on $m_l$ will have the form $\mathcal{N}(\mu_l, c_l)$.
Updating this posterior with a single embedding $e_i$ has the form:
\begin{align*}
    m &\gets \frac{c_l}{c_l + 1} m + \frac{1}{c_l + 1} e_i \\
    c_l &\gets c_l + 1
\end{align*}
This is precisely the update in Algorithm~\ref{alg:clustering}.

Note that in this model, the counts $c_l$ are also the precision parameters of the distribution, representing the inverse spread (or the concentration) of each cluster.
At each step of Algorithm~\ref{alg:clustering}, these counts are decayed.
Effectively, this causes the variance of the distribution representing each cluster to spread out: thus at each time step, each cluster becomes less concentrated and more uncertain about which data points belong to it.
The hyperparameter $\gamma$ captures the rate of diffusion of all clusters in this manner. This uncertainty increase applied at each step acts as a ``forgetting" mechanism that helps the algorithm to deal with a changing data distribution. 

Cluster re-sampling, as already justified for $e_t$ above in terms of \eqref{eq:stepwise}, ensures that the number of clusters is bounded by $|M|$.
There are two details to examine: what is merged, and how it is merged.
As $c_j \mapsto 0$, the probability assigned by the Gaussian likelihood of cluster $j$ to any new datum approaches zero also, thus the cluster with the lowest counts is likely to have the least impact on future density estimates (as it is most diffuse). 
When $c_j \gg 0$, however, it is not so clear which cluster should be removed. Therefore, we stochastically select which cluster to remove with probability inversely proportional to the square of the counts (using the square of the counts emphasizes small differences in counts more than $1/c_j$).
The cluster could potentially be removed completely, but we instead choose to re-assign its counts to the nearest cluster as we experimentally found this strategy to be less sensitive to the choice of hyperparameters. 

\begin{figure}[H]
\centerline{
\includegraphics[width=.85\textwidth]{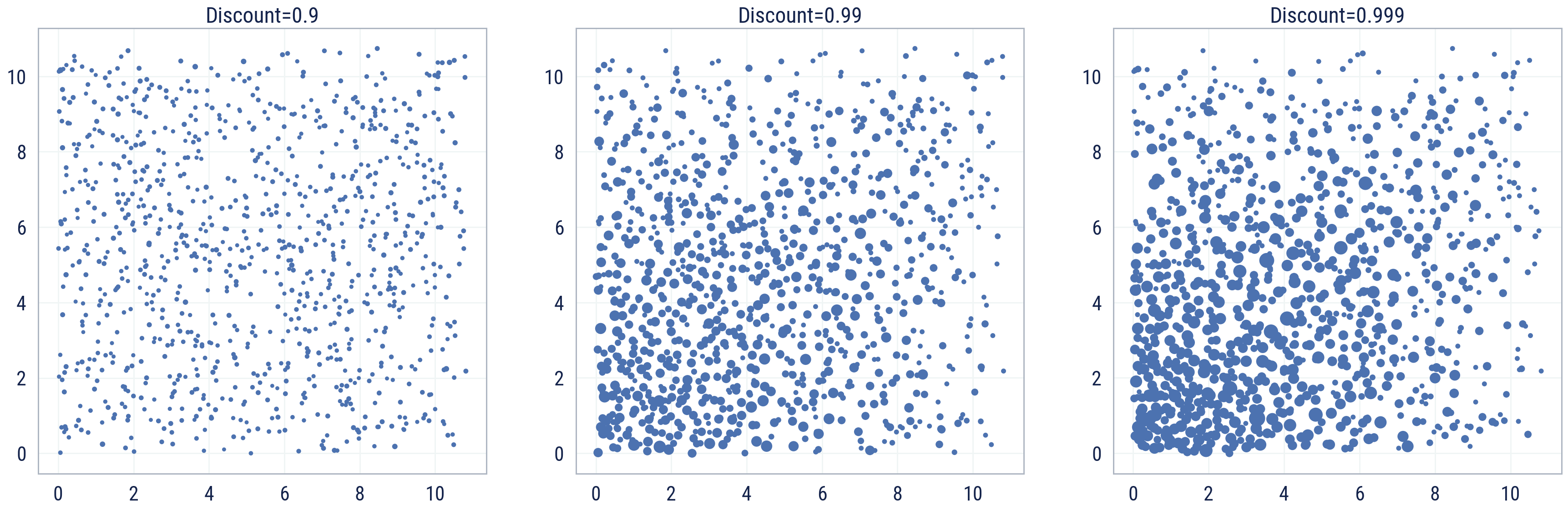}
}
\caption{Non-stationary density estimation using \DETOCS on a toy example. For step $t=0,\ldots, 100$, we sample a batch of $64$ 2D-embeddings uniformly from the square of side $1 + \sqrt{t}$. The support of the embedding distribution therefore expands over time to simulate a non-stationary distribution similar to the distribution of states visited by an RL agent over the course of exploration. 
We plot the atoms learned by \DETOCS with a size proportional to their count. We find that for a small enough discount, \DETOCS exhibits a short-term memory, accurately approximating the distribution of the final distribution. As we increase the discount, \DETOCS exhibits a longer-term memory, approximating the historical density of states, as can be seen by the concentration of probability mass in the bottom-left~corner.}  \label{fig:devo_density_estimation_toy}
\end{figure}

To help build some intuition about the effects of the discount factor, we illustrate its effects on a toy example with a non-stationary embedding distribution in Fig.~\ref{fig:devo_density_estimation_toy}.
We find that tuning the discount $\gamma$ allows to smoothly interpolate between short-term and long-term memory.

\begin{figure}[H]
\centerline{
\includegraphics[width=.9\textwidth]{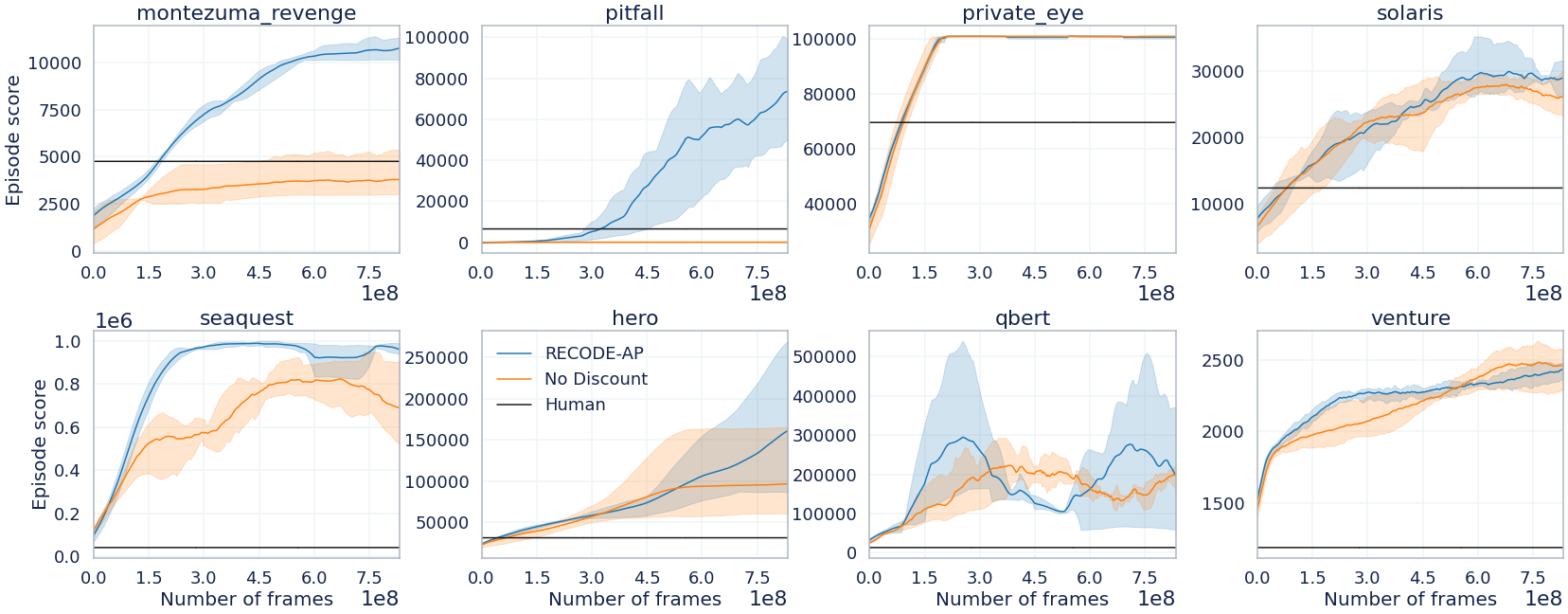}
}
\caption{Effect of discount on performance. As embeddings evolve throughout the training it may happen that older clusters stop being meaningful under the current representation. Deactivating the discount (i.e. $\gamma=1$) results in a significant degradation in performance, especially in hard-exploration settings like \emph{Montezuma's Revenge} and \emph{Pitfall!}.}
\label{fig:no_discount_atari}
\end{figure}

\begin{figure}[H]
\centerline{
\includegraphics[width=.45\textwidth]{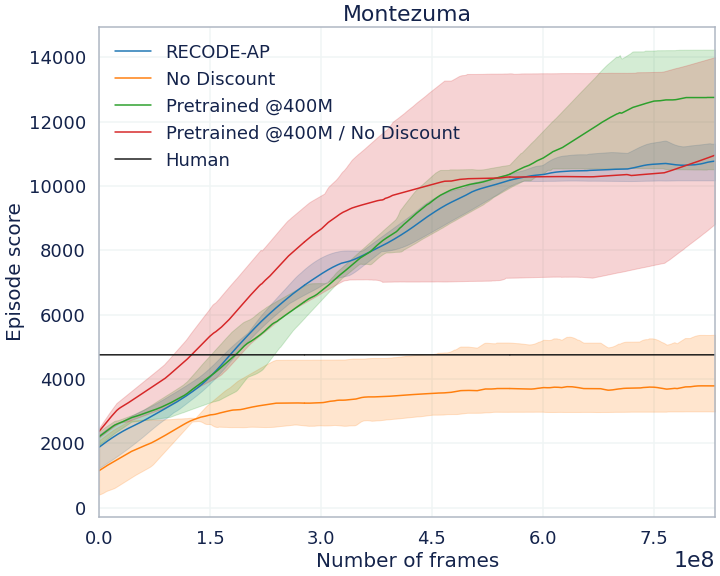}
}
\caption{Pretrained vs concurrently-trained embeddings and sensitivity to discount. 
We take a snapshot of the embeddings at 400M frames and then the agent is trained again from scratch with these frozen embeddings. 
We find that we can achieve similar performance even with a $\gamma=1$ (i.e. no discount). 
Interestingly, we also observe that the agent can achieve much higher scores than those the original agent had achieved at the time the snapshot was taken. }\label{fig:pretrained_montezuma}
\end{figure}

To confirm the practical necessity of cluster discount, we perform additional ablations on the Atari environment. 
If we don't train the representation during the exploration, but start with a pretrained one, we see that \DETOCS can perform reasonably well also without using discount (see Fig.~\ref{fig:pretrained_montezuma}).
However, as shown in Fig.~\ref{fig:no_discount_atari}, as soon as we also train the representation at the same time, the ability to forget old observations allows to compensate distribution-shift and achieves quite better performance.

 \section{Analyzing Exploration with \DETOCS}
\label{appendix:toy_example}

\begin{figure}[H]

    \begin{subfigure}{.35\textwidth}
        \centering
        \includegraphics[width=0.8\textwidth]{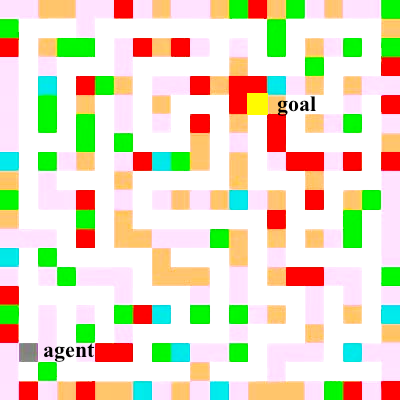}
    \end{subfigure}
    \begin{subfigure}{.65\textwidth}
        \centering
        \includegraphics[width=0.8\textwidth]{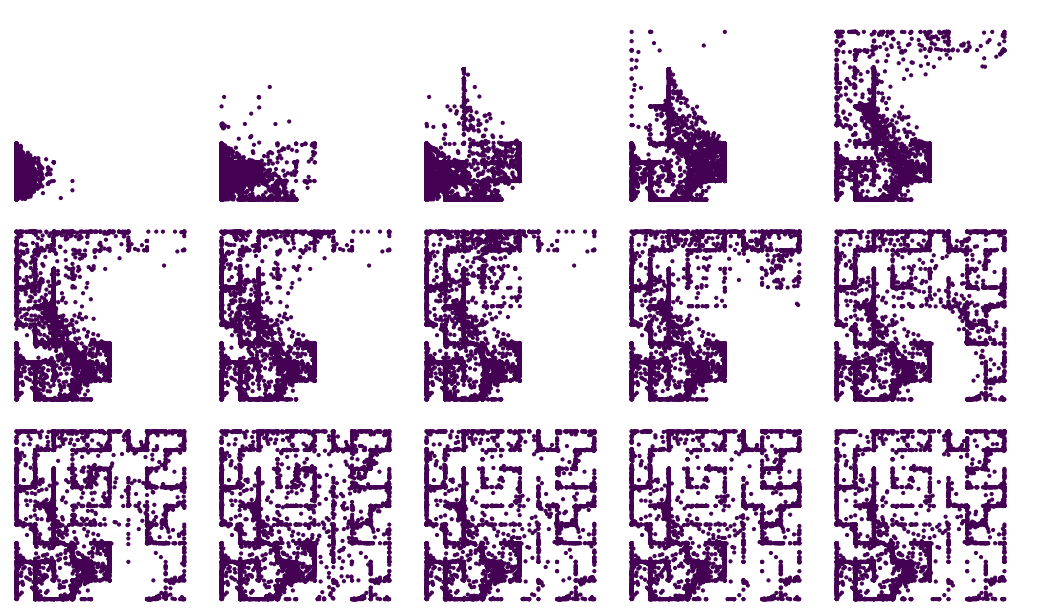}
    \end{subfigure}  
    \caption{(Left) Random Disco Maze (Right) Evolution of the distribution of the clusters learned by \DETOCS over time, see text for details.}
    \label{fig:coverage_toy_example}
    
\end{figure}

In this section, we present a simple example to show a simple example to illustrate how the exploration process unfolds for \DETOCS. We use a variant of the \emph{Random Disco Maze}: a grid-world environment proposed in \citealp{badia2020never} to show the the importance of estimating the exploration bonus using a controllable state representation, depicted in Fig.~\ref{fig:coverage_toy_example} (left). The agent starts each episode in a fixed position of a fully observable maze of size 21x21. The agent can take four actions \{left, right, up, down\}. The episode ends if the agent steps into a wall, reaches the goal state or reaches a maximum of 500 steps.

Crucially the environment presents random variation in the color of each wall fragment at every time step. Specifically, the color of each wall fragment is randomly and independently selected from a set of five possible colors. This introduces a great deal of irrelevant variability into the system, which presents a serious challenge to exploration bonus methods based on novelty. The reason for this is that the agent will never see the same exact state twice, as the colors of the wall fragments will always be different each time step.

We ran \DETOCS for 100 million steps. The agent is able to find the goal after collecting around 50 to 60 million steps. In Fig.~\ref{fig:coverage_toy_example} (right) shows how the distribution of clusters changes as the agent explore this environment. As the agent always starts in the same position (bottom-left corner of the maze), the distribution is heavily skewed towards over-representing this points. We can see that as time progresses the cluster centers uniformly cover all the maze.

\begin{figure}[H]
    \centering
    \includegraphics[width=0.45\textwidth]{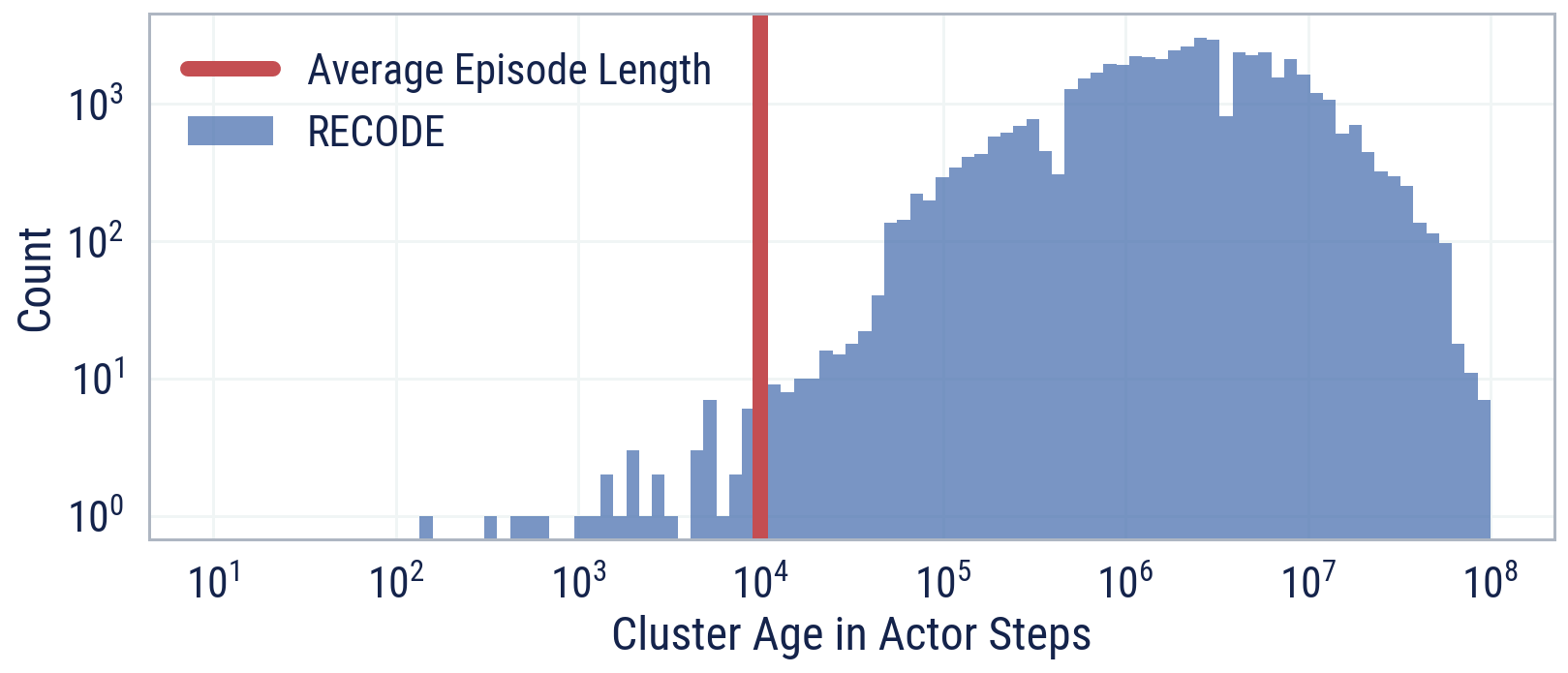}
    \caption{Age distribution of the clusters learned by \DETOCS on \emph{Montezuma's Revenge}. \DETOCS's memory horizon spans much more than a single episode. 
    We set $\gamma = 0.999$ as in the experiments of Fig.~\ref{fig:detocs_atari}. 
    We indicate in red the average length of an episode, showing that in this setting, \DETOCS's memory reaches back thousands of episodes.}
    \label{fig:devo_cluster_ages}
\end{figure}

We investigate how far back the memory of \DETOCS goes in \emph{Montezuma's Revenge}. The results are shown in Fig.~\ref{fig:devo_cluster_ages}. 
We find that the distribution of the age of the clusters learned by \DETOCS (i.e. how many steps ago each atom has been inserted in the memory) exhibits a mode around $2\cdot 10^6$ actor steps, which corresponds to hundreds of episodes, with a significant number of clusters ten times older than that. We remind that \NGU's short-term non-parametric novelty estimated at most one episode (red line in the Figure).

 \section{Hyper-parameters}
\label{appendix:hypers}
Here we show the precise hyper-parameter values used in our experiment, Table~\ref{tab:agent_hypers_atari} for Atari and Table~\ref{tab:agent_hypers_dmh} for \DMH
We omit hypers which do not differ from the base agent, MEME~\cite{kapturowski2022human}.

\begin{table}[H]
    \centering
    \caption{Atari Hyper-parameters.}
    \label{tab:agent_hypers_atari}
    \resizebox{0.55\textwidth}{!}
    {
        \begin{tabular}{lrrr}
        \toprule
        \textbf{Parameter} & \textbf{Value} \\
        \midrule
      
        \DETOCS memory size & $5 \times 10^4$\\
        \DETOCS discount $\gamma$ & $0.999$ \\
        \DETOCS insertion probability $\eta$ & $0.05$ \\
        \DETOCS relative tolerance $\kappa$ & $0.2$ \\
        \DETOCS reward constant $c$ & $0.01$ \\
        \DETOCS decay rate $\tau$ & $0.9999$ \\
        \DETOCS neighbors $k$ & $20$ \\
        IM Reward Scale $\beta_{\mathrm{IM}}$ & $1.0$ \\

        Max Discount & $0.9997$ \\
        Min Discount & $0.97$ \\
        Replay Period & $80$ \\
        Trace Length & $160$ \\
 
        Replay Ratio & $6.0$ \\
        Replay Capacity & $2 \times 10^{5}$ trajectories \\
        
Batch Size & $64$ \\
        RL Adam Learning Rate & $3 \times 10^{-4}$ \\
        Emedding Adam Learning Rate & $6 \times 10^{-4}$ \\
        RL Weight Decay & $0.05$ \\
        Embedding Weight Decay & $0.05$ \\
        
        RL Torso initial stride & $4$ \\
        RL Torso num blocks & $(2, 3, 4, 4)$ \\
        RL Torso num channels & $(64, 128, 128, 64)$ \\
        RL Torso strides & $(1, 2, 2, 2)$ \\

        \bottomrule
        \end{tabular}
    }
\end{table}

\begin{table}[H]
    \centering
    \caption{\DMH Hyper-parameters.}
    \label{tab:agent_hypers_dmh}
    \resizebox{0.55\textwidth}{!}
    {
        \begin{tabular}{lrrr}
        \toprule
        \textbf{Parameter} & \textbf{Value} \\
        \midrule
      
        \DETOCS memory size & $2 \times 10^5$\\
        \DETOCS discount $\gamma$ & $0.997$ \\
        \DETOCS insertion probability $\eta$ & $0.2$ \\
        \DETOCS relative tolerance $\kappa$ & $0.2$ \\
        \DETOCS reward constant $c$ & $0.01$ \\
        \DETOCS decay rate $\tau$ & $0.9999$ \\
        \DETOCS neighbors $k$ & $20$ \\
        IM Reward Scale $\beta_{\mathrm{IM}}$ & $0.1$ \\

        Max Discount & $0.997$ \\
        Min Discount & $0.97$ \\
        Replay Period & $40$ \\
        Trace Length & $80$ \\
 
        Replay Ratio & $2.0$ \\
        Replay Capacity & $5000$ trajectories \\
        
Batch Size & $128$ \\
        RL Adam Learning Rate & $1 \times 10^{-4}$ \\
        Embedding Adam Learning Rate & $3 \times 10^{-4}$ \\
        RL Weight Decay & $0.1$ \\
        Embedding Weight Decay & $0.1$ \\
        
        RL Torso initial stride & $2$ \\
        RL Torso num blocks & $(2, 4, 12, 6)$ \\
        RL Torso num channels & $(64, 128, 128, 64)$ \\
        RL Torso strides & $(1, 2, 2, 2)$ \\

        \bottomrule
        \end{tabular}
    }
\end{table}

\begin{table}[H]
    \centering
    \caption{\CASM Hyper-parameters.}
    \label{tab:casm_hypers_dmh}
{
        \begin{tabular}{lrrr}
        \toprule
        \textbf{Parameter} & \textbf{Value} \\
        \midrule
        
         Transformer Type & GatedTransformerXL \\
         State Mask Rate & 0.8 \\
         Num Masks Per Trajectory & 4 \\
         Action Embedding Size & 32 \\
         Num Layers & 2 \\
         Attention Size & 128 \\
         Num Attention Heads & 4 \\
         MLP Hidden Sizes & (512,) \\
         Predictor Hidden Sizes & (128,) \\

        \bottomrule
        \end{tabular}
    }
\end{table}

 \section{Architecture of a distributed agent using \DETOCS}
\label{appendix:architecture}
\begin{figure}[H]
\centerline{
\includegraphics[width=.7\textwidth]{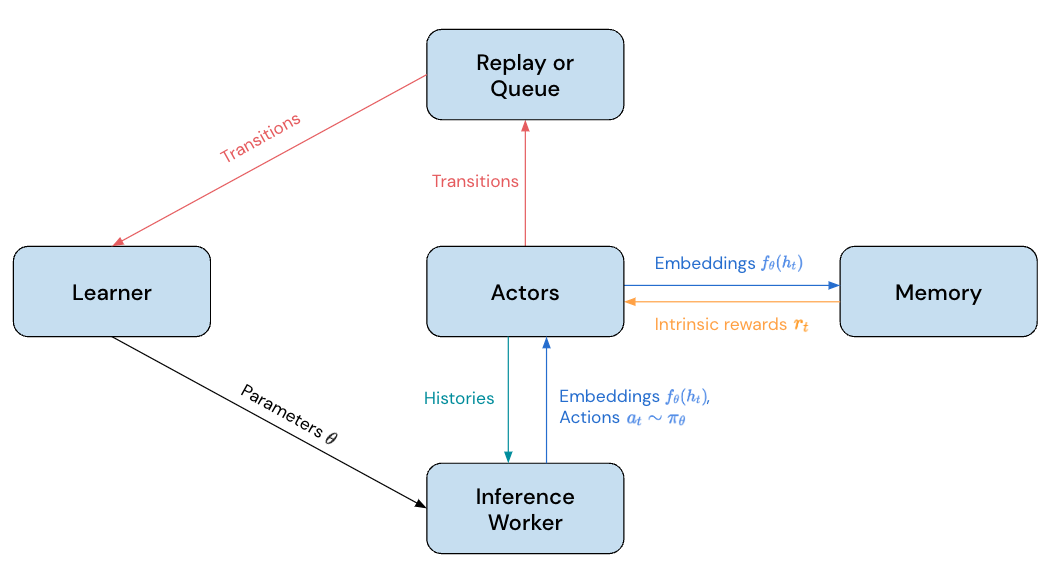}
}
\caption{Overview of the architecture of a distributed agent using \DETOCS.}\label{fig:detocs_architecture}
\end{figure}

We now detail how \DETOCS can be efficiently integrated in a typical distributed RL agent \citep{espeholt2018impala, kapturowski2018recurrent} that comprises several processes that run in parallel and interact with each other, allowing for large-scale experiments.
Classically, a Learner performs gradient steps to train a policy $\pi_\theta$ and an embedding (representation) function $f_\theta$, forwarding the parameters $\theta$ to an Inference Worker. 
A collection of independent Actors query the inference worker for actions that they execute in the environment and send the resulting transitions to the Learner, optionally through a (prioritized) Replay \citep{mnih2015human,PrioritizedReplay}. 
When using \DETOCS, the Actors additionally communicate with a shared Memory implementing Algorithm~\ref{alg:recode}: at each step $t$, they query from the Inference Server an embedding $f_\theta(h_t)$ of their history and send it to the shared Memory which returns an intrinsic reward $r_t$ that is then added to the extrinsic reward to train the policy in the Learner process. 
A diagram giving an overview of the typical architecture of a distributed agent using \DETOCS is given in  Figure~\ref{fig:detocs_architecture}.

 \section{Agent Taxonomy}
\label{appendix: taxonomy}

All methods evaluated in the experiments in the main paper and the extended experiments in App.~\ref{appendix:ablations} are composed by three main components.

\begin{itemize}
\itemsep0em 
\item A base agent that oversees the overall RL learning process (e.g., executing actions and collection observations, computing adjusted returns, updating the policy, ...). We focus on \MEME~ \citep{kapturowski2022human}, a recent improvement over Agent57~\citep{badia2020agent57} that achieves much greater sample efficiency and is the current state-of-the-art on Atari, and a VMPO-based agent \citep{guo2022byol} that is the current state-of-the-art on \DMH.
\item A representation learning mechanism to generate observation embeddings which are fed to the intrinsic reward generator. We consider both Action Prediction (\AP) and \CASM embeddings. Note that some intrinsic reward modules cannot make effective use of the representation learning module (e.g., \RND), while others merge both second and third module in a single approach (e.g., \BYOLE)
\item An algorithm to generate intrinsic rewards. In addition to \DETOCS, we also consider the recent \BYOLE \cite{guo2022byol}, \NGU \cite{badia2020never} and \NGU's two building blocks, \RND \cite{burda2019exploration} and Episodic Memory (\EMM) \cite{pritzel2017neural}.
\end{itemize}

For example, in our more detailed taxonomy the original \MEME agent described in \cite{kapturowski2022human} is denoted as the \MEME-\NGU-\AP baseline, and compared against our novel \MEME-\DETOCS-\AP agent where the only modifications is the changed exploration reward. Table \ref{appendix:ablations} reports more details on all combinations available present in our experiments.
\begin{table}[H]
    \centering
    \caption{Taxonomy of agents used in the experiments.}
    \label{tab:tab:taxonomy-algs}
        \resizebox{\textwidth}{!}{
        \begin{tabular}{lllll}
        \toprule
        \multicolumn{2}{l}{\textbf{Agent name}} & \textbf{Base agent} & \textbf{Intrinsic reward} & \textbf{Representation learning}\\
        \midrule
        \MEME-\NGU-\AP &\cite{kapturowski2022human}  & \MEME & \NGU & \AP\\
        \MEME-\RND &(ablation)  & \MEME & \RND & N/A\textsuperscript{(a)}\\
        \MEME-\EMM-\AP &(ablation)  & \MEME & \EMM & \AP\\
        \MEME-\texttt{RNDonAP} &(ablation)  & \MEME & \RND & \AP\textsuperscript{(b)}\\
        \MEME-\DETOCS-\AP &(this paper)  & \MEME & \DETOCS & \AP\\
        \MEME-\DETOCS-\CASM &(this paper)  & \MEME & \DETOCS & \CASM\\
        \MEME-\NGU-\CASM &(ablation)  & \MEME & \NGU & \CASM\\
        \VMPO-\BYOLE &\cite{guo2022byol}  & \VMPO & \BYOLE & \BYOLE\textsuperscript{(c)}\\
        \midrule
        \multicolumn{5}{l}{
         (a) {As in the original paper \RND takes as input raw observations}.}\\[.1cm]
        \multicolumn{5}{l}{
        \begin{minipage}{16cm}
        (b) To test \RND's ability to cope with non-stationary representations, we train an \AP encoder concurrently with the policy and use it to 
        create embeddings of the observations that are fed in \RND (i.e., running 
        \RND on top of \AP).
        \end{minipage}
        }\\[.3cm]
        \multicolumn{5}{l}{
        \begin{minipage}{16cm}
         (c) The \BYOLE mechanism internally trains a neural network to 
         predict the dynamical evolution of the observations. This provides the agent with both a reward/novelty signal (prediction error) as well as an embedded representation of the observations (that can be extracted from the last few layers of the 
         network).
         \end{minipage}
         }\\
        \bottomrule
        \end{tabular}
        }
\end{table}

\section{Exploration in the \DMH environment}
\begin{figure}[H]
\centerline{
\includegraphics[width=.9\textwidth]{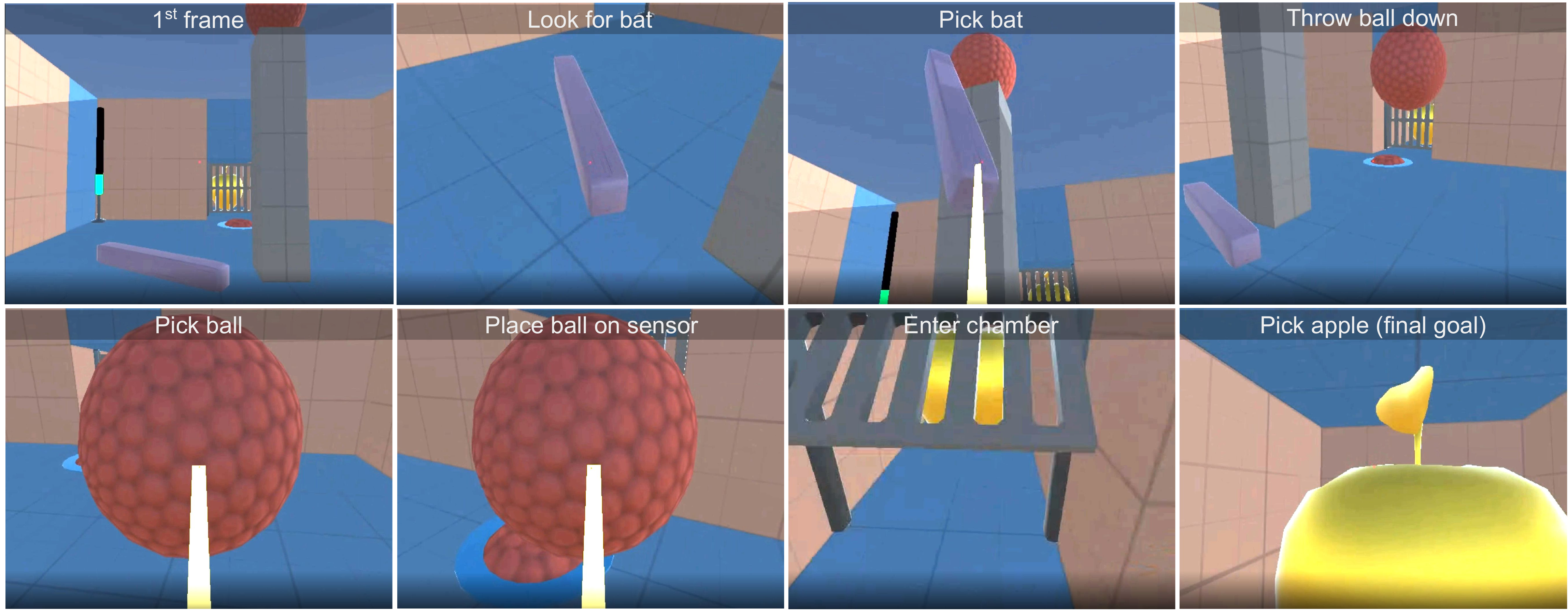}
}
\caption{First-person-view snapshots of an agent solving the \DMH \texttt{Baseball} task. Images are ordered chronologically from left to right and top to bottom. Each image depicts a specific stage of the task. The agent must interact with specific objects in the environment in order to solve the task.}\label{fig:agi_human_knock_obj_demo}
\end{figure}

The three-dimensional tasks in \DMH can have an extremely large state space to explore. Consider for example the \texttt{Baseball task}, as shown in Fig. \ref{fig:agi_human_knock_obj_demo}: the agent needs to look at the scene, find the bat, pick it up, throw the ball down, pick up the ball and be able to get the apple.

 \section{Aggregated Results}
\label{appendix:aggregated_results}

In this section, we show the aggregated results over all different environments, both for the Atari suite and for \DMH.
To ensure that no single environment dominates due to larger reward scales we use the Human Normalized Score~\citep{mnih2015human} in each environment, and then cap scores above 100\% prior to averaging. 

\begin{figure}[H]

    \begin{subfigure}{.5\textwidth}
        \centering
        \includegraphics[width=0.8\textwidth]{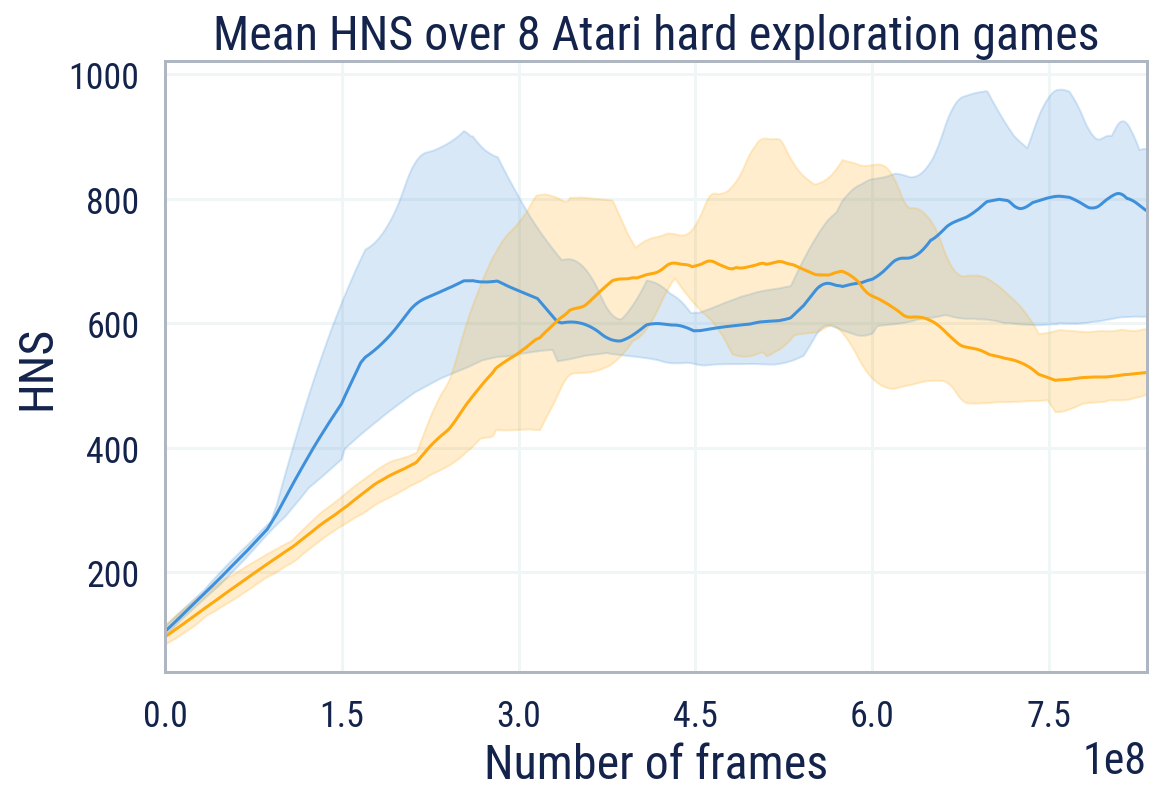}
    \end{subfigure}
    \begin{subfigure}{.5\textwidth}
        \centering
        \includegraphics[width=0.8\textwidth]{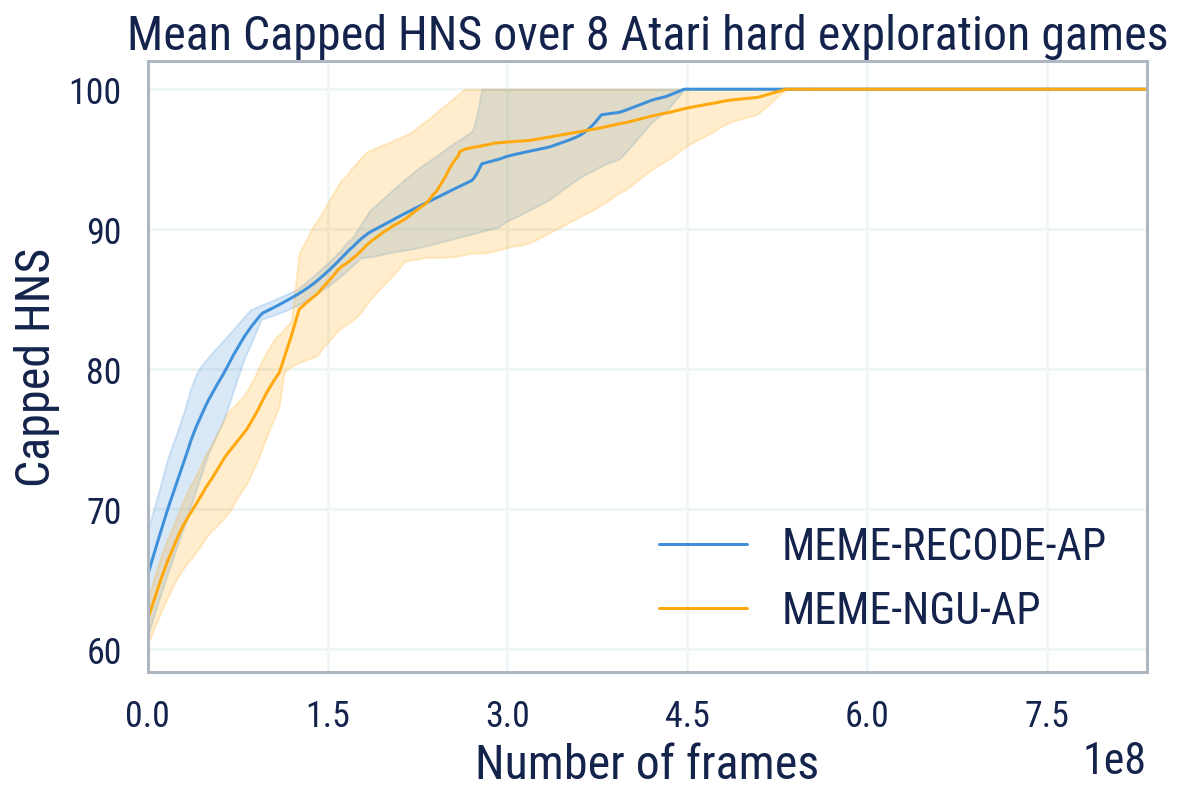}
    \end{subfigure}  
    \caption{Aggregated results on Atari. 
    (Left): Mean Human-Normalized Scores of \MEME-\DETOCS-\AP compared to \MEME-\NGU-\AP on Atari games. 
    (Right): Capped Human-Normalized Scores.}
    \label{fig:mean_capped_atari}
    \centering
\end{figure}

As Fig.~\ref{fig:mean_capped_atari} shows, (Left), the uncapped score can swing significantly over time, which in this case is simply an artifact the high variance present in \textit{Q*bert}. This variance arises due to a bug in \textit{Q*bert}, which allows for much larger scores to be obtained if exploited.

\begin{figure}[H]
    \centering
    \includegraphics[width=.45\textwidth]{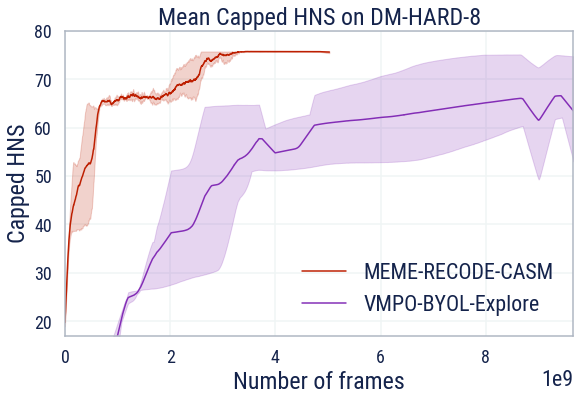}
    \caption{Aggregated results on \DMH.
    Mean Capped Human-Normalized Scores of \MEME-\DETOCS-\CASM are compared to \VMPO-\BYOLE.}
    \label{fig:mean_capped_hard8}
\end{figure}

In Fig~\ref{fig:mean_capped_hard8}, we present our main results aggregated over all environments in each \DMH task suite. Table~\ref{tab:aggregated-results-dmh} summarizes the results and compares the performance of the \DETOCS novelty reward mechanism with that of \NGU. We emphasize how effective our approach is when applied to three-dimensional environments like \DMH, if compared to alternatives like BYOL-Explore.

\begin{table}[H]
    \caption{Aggregated results over \DMH environment tasks.}
    \label{tab:aggregated-results-dmh}
    \centering
    \resizebox{\textwidth}{!}
    {
    \begin{tabular}{llllll}
    \toprule
    \textbf{Game} & \textbf{Human} & \textbf{\MEME-\NGU-\AP} & \textbf{\MEME-\NGU-\CASM} & \textbf{\MEME-\DETOCS-\AP} & \textbf{\MEME-\DETOCS-\CASM}\\
    \midrule
 Baseball & 7.90 & 0.00 $\pm$ 0.00 & 0.00 $\pm$ 0.00 &  \bf{10.00 $\pm$ 0.00} & \bf{10.00 $\pm$ 0.00} \\
 Drawbridge & 12.30 & 0.00 $\pm$ 0.00 & 4.15 $\pm$ 6.62 & 12.41 $\pm$ 0.07 & \bf{12.86 $\pm$ 0.13} \\
 Navigate Cubes & 7.80 & 3.33 $\pm$ 5.33 & 9.76 $\pm$ 0.25 &  \bf{10.00 $\pm$ 0.00} & \bf{10.00 $\pm$ 0.00} \\
 Push Blocks & \bf{8.40} & 0.00 $\pm$ 0.00 & 0.24 $\pm$ 0.24 & 1.07 $\pm$ 0.62 & 4.06 $\pm$ 2.14 \\
 Remember Sensor & \bf{7.60} & 0.00 $\pm$ 0.00 & 0.00 $\pm$ 0.00 & 0.00 $\pm$ 0.00 & 0.00 $\pm$ 0.00 \\
 Throw Across & 5.70 & 0.00 $\pm$ 0.00 & 0.00 $\pm$ 0.00 & 0.00 $\pm$ 0.00 & \bf{9.72 $\pm$ 0.31} \\
 Wall Sensor & 9.10 & 0.00 $\pm$ 0.00 & 0.00 $\pm$ 0.00 & \bf{10.00 $\pm$ 0.00} &  \bf{10.00 $\pm$ 0.00} \\
 Wall Sensor Stack & \bf{8.60} & 0.00 $\pm$ 0.00 & 0.00 $\pm$ 0.00 & 0.00 $\pm$ 0.00 & 0.00 $\pm$ 0.00 \\
    \bottomrule
    
    \end{tabular}
    }

\end{table}

 \section{Multitask Experiments}
\label{appendix:multitask}

\begin{figure}[H]
    \centering
    \includegraphics[width=0.99\textwidth]{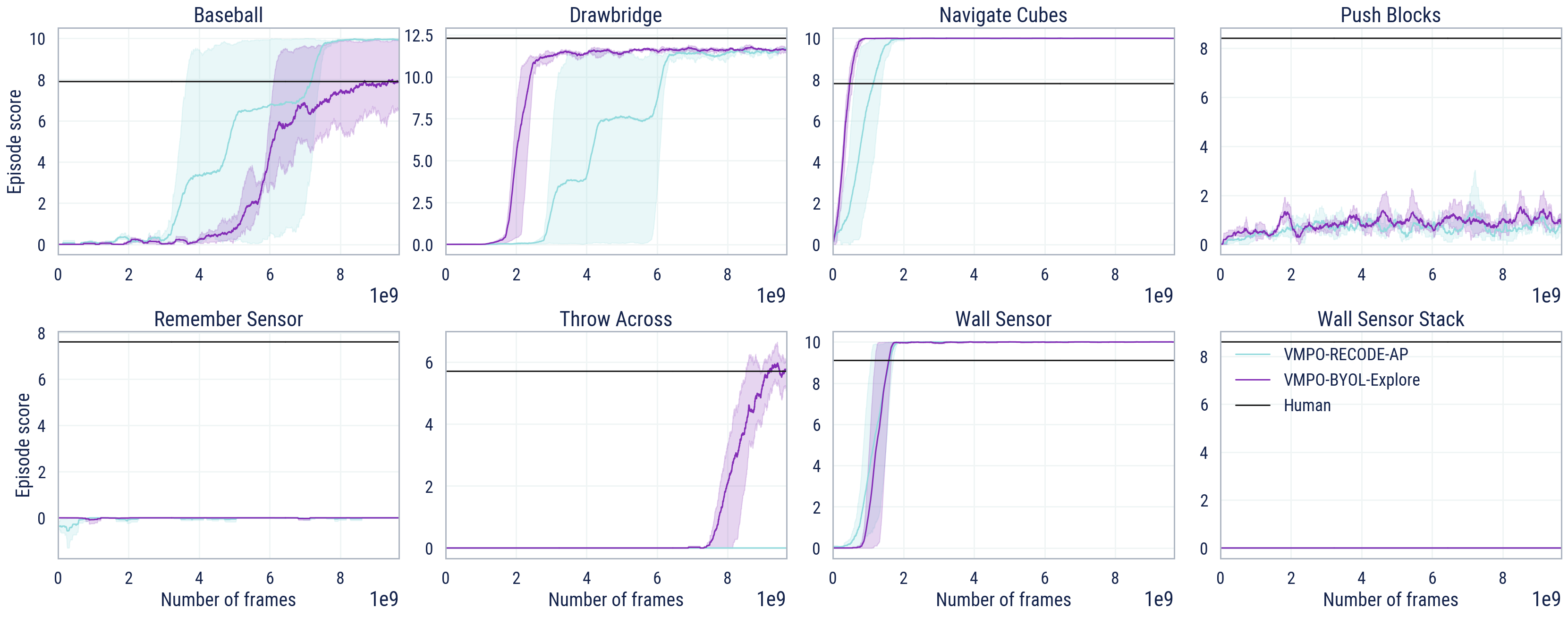}
    \caption{Performance of \DETOCS compared to BYOL-Explore on the multi-task version of \DMH. Our \DETOCS implementation in this experiments is based on VMPO, using a continuous action set.\label{fig:dm_hard_8_multi_task}}
\end{figure}

We also implemented \DETOCS in a VMPO-based agent similar to the one used with BYOL-Explore~\citep{guo2022byol}, and compared our performance with BYOL-Explore in the multi-task setting. This experiment serves two different purposes. 
First, this demonstrates the generality of our exploration bonus, that is shown to be useful in widely different RL agents, be they value-based or policy-based. 
Second, we can do a direct comparison with the state of the art BYOL-Explore agent in the multi-task settings. 
However, we note that the representation learning technique used in this experiment, 1-step Action Prediction, is based on a feed-forward embedding that discards past history, and may therefore not be the best fit for exploration in Partially Observable MDPs (POMDPs). 
Still, Fig.~\ref{fig:dm_hard_8_multi_task} shows that \DETOCS's performance is competitive with that of BYOL-Explore, with only one level missing to match its performance. 
Improving this performance using better-suited representations, such as \CASM, is left for future work. 

 \section{Additional ablation studies}
\label{appendix:ablations}

\subsection{CASM masking}
Here we analyze the performance of our technique when removing the masking strategy in \CASM (see main text). For each element of the sequence, instead of providing either the embedded observation or the action, we always provide both, making the classifier upstream task too simple. Masking allows to provide extra context (with respect to action-prediction) while keeping the prediction task hard enough to require the encoding of high-level features in the representation. 

\begin{figure}[H]

    \begin{subfigure}{.5\textwidth}
        \centering
        \includegraphics[width=0.8\textwidth]{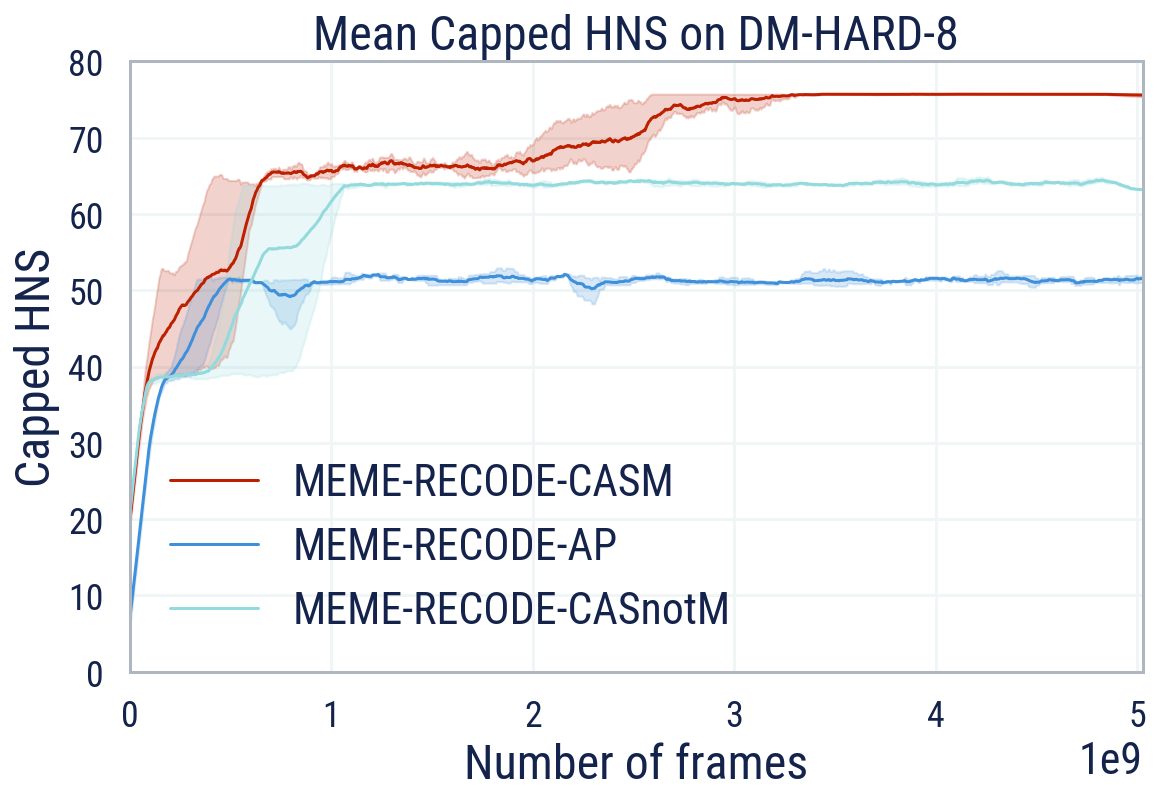}
    \end{subfigure}
    \begin{subfigure}{.5\textwidth}
        \centering
        \includegraphics[width=0.8\textwidth]{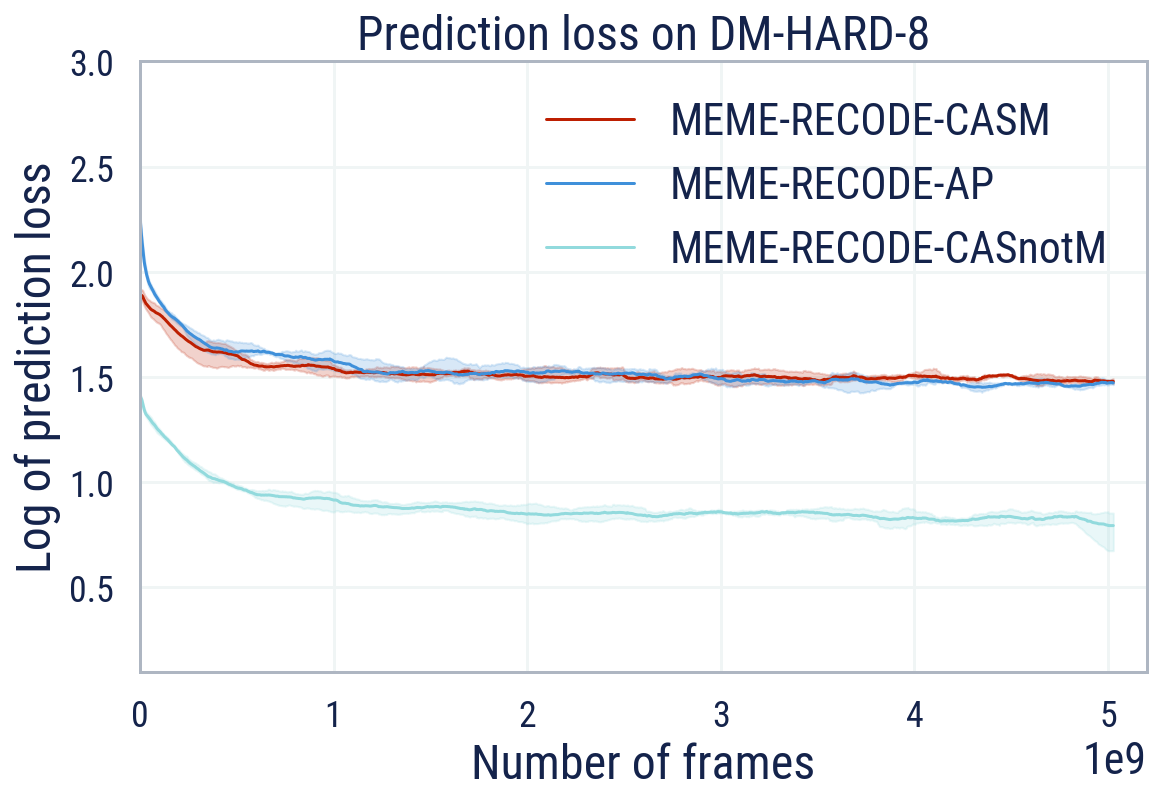}
    \end{subfigure}  
    \caption{\CASM masking ablation. (Left):  \CASnotM is \CASM without any masking applied to the trajectory. Even by itself, the extra context provides a clear advantage over \AP, but the masking strategy is essential to solve Push Block. (Right): The loss for \CASnotM is much lower than for both \AP and \CASM suggesting that the additional context without any masking makes the prediction task easier (but leading to a less robust representation).}
    \label{fig:casm-masking}
    
\end{figure}

\subsection{\RND on top of Action Prediction Embeddings}

\begin{figure}[H]
    \centering
    \includegraphics[width=\textwidth]{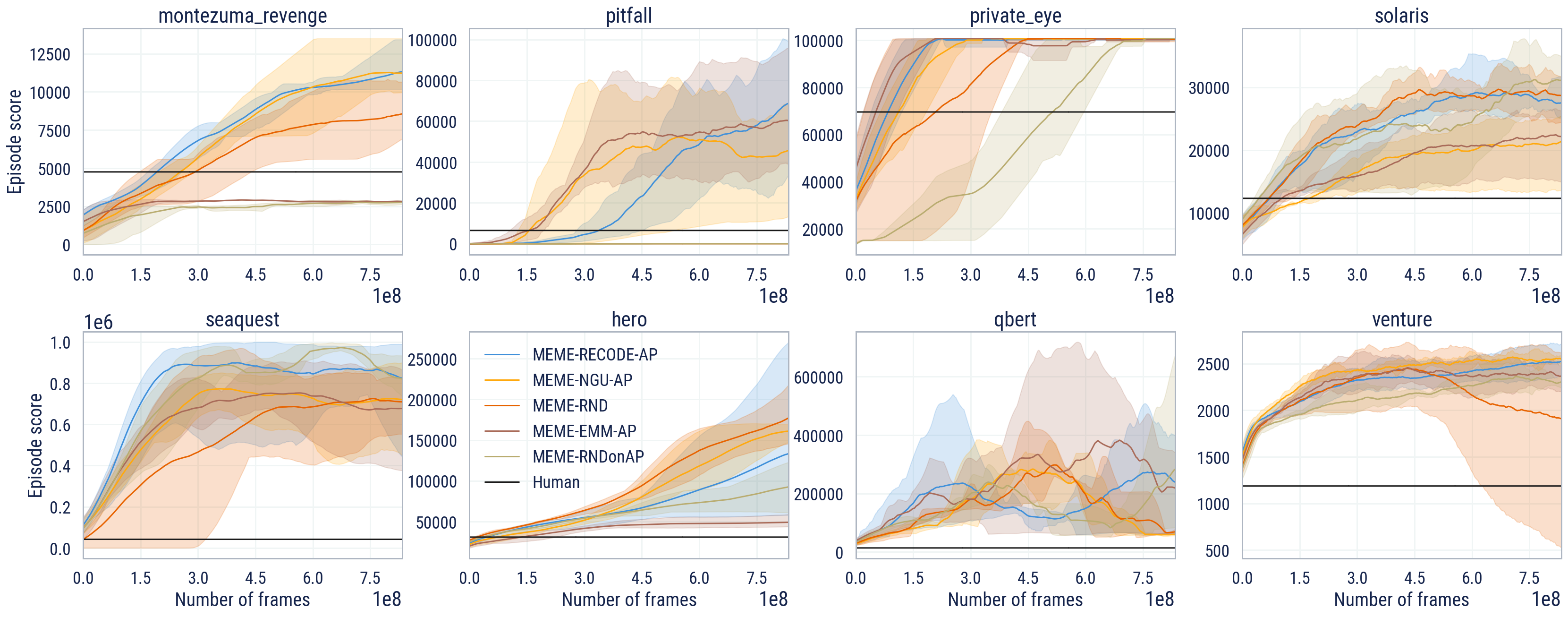}
    \caption{Performance of \DETOCS compared to \MEME and its ablations on 8 hard exploration Atari games. We find that this approach does not allow to solve some of the hardest games such as \textit{Montezuma's Revenge} or \textit{Pitfall!}}
    \label{fig:npde_atari_all}
\end{figure}

We adapt \RND to leverage trained action-prediction embeddings, which we refer to as \RNDonAP. To that effect, we use a randomly initialized Multi-Layer Perceptron (MLP) to perform a random projection of the embedding, and use a second, trained MLP, to reconstruct this random projection. The reconstruction error provides an intrinsic reward for exploration, which we normalize by a running estimate if its standard deviation as in \cite{burda2019exploration}. We find that the resulting agent is unable to solve some of the hardest exploration games such as \texttt{Montezuma's Revenge} or \texttt{Pitfall}.
The results of this ablation is shown in Fig.~\ref{fig:npde_atari_all}. Experiments with pre-trained embeddings do seem to indicate that \RNDonAP can obtain stronger performance in this setting, but the inability to concurrently train the embeddings greatly limits the general applicability of the method.

One possible explanation of this failure is the fact that a large \RND error can be caused by either the observation of a new state, or a drift in the representation of an already observed one.
The failure of \RND to disentangle these two effects results in poor exploration.

\subsection{\DETOCS on top of BYOL embeddings}

\begin{figure}[H]
\centerline{
\includegraphics[width=.9\textwidth]{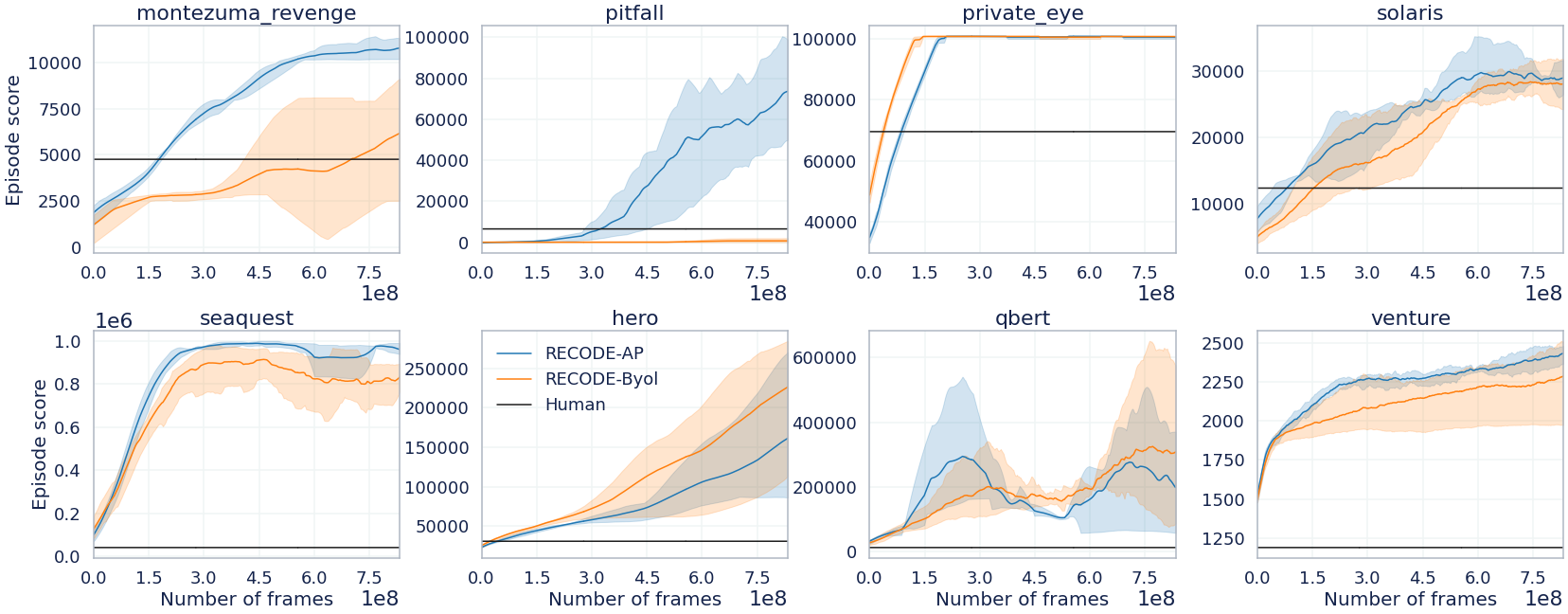}
}
\caption{Comparison of \DETOCS with AP vs BYOL embeddings. We observe that \DETOCS is able to leverage BYOL embeddings to achieve superhuman performance on \emph{Montezuma's Revenge} and achieve positive scores on \emph{Pitfall!}, though significantly underperforming compared to AP embeddings. For BYOL, we sweeped over embedding sizes of $\{32, 128, 512\}$ and report the best performing size, $32$.}\label{fig:recode_byol_atari}
\end{figure}

To further show \DETOCS robustness to change of representation, in Fig.~\ref{fig:recode_byol_atari}, we compare the performance of the \DETOCS embedding on top of an action-prediction representation with respect to a BYOL representation (as in ~\citet{guo2022byol}).

\end{document}